\documentclass[conference]{IEEEtran}
\IEEEoverridecommandlockouts
\usepackage{graphicx}
\usepackage{amsmath, amssymb, amsthm}
\usepackage{lipsum}  %
\usepackage{xcolor}
\usepackage{caption}
\usepackage{subcaption}
\usepackage{booktabs} 
\usepackage{hyperref} 
\usepackage{csquotes} 
\usepackage{array}  %
\usepackage{multirow} 

\newcommand{\customsubcap}[1]{\caption{\fontsize{8}{9}\selectfont #1}}
\newcommand{\figref}[1]{Figure~\ref{#1}}
\newcommand{\tabref}[1]{Table~\ref{#1}}
\newcommand{\secref}[1]{Section~\ref{#1}}

\renewcommand{\vec}[1]{\boldsymbol{#1}}

\newcommand{\thre}{\text{th}} 
\newcommand{\subfire}{\text{f}}
\newcommand{\supinit}{\text{init}}
\newcommand{\gt}{\text{gt}}
\newcommand{\pgt}{\text{pgt}}
\usepackage{cite}
\usepackage{amsmath,amssymb,amsfonts}
\usepackage{algorithmic}
\usepackage{graphicx}
\usepackage{textcomp}
\usepackage{xcolor}
\def\BibTeX{{\rm B\kern-.05em{\sc i\kern-.025em b}\kern-.08em
    T\kern-.1667em\lower.7ex\hbox{E}\kern-.125emX}}

\let\originalchi\chi
\renewcommand{\chi}{{\mathpalette\raisechi{}}}
\newcommand{\raisechi}[2]{%
  \raisebox{0.3ex}{$#1\originalchi$}%
}

\begin{document}

\title{Prediction of the Most Fire-Sensitive Point in Building Structures with Differentiable Agents for Thermal Simulators}

\author{
    \IEEEauthorblockN{Yuan Xinjie}
    \IEEEauthorblockA{Shenzhen International Graduate School\\
        Tsinghua University\\
        Shenzhen, Guangdong, China \&\\
        Pacific Earthquake Engineering Research (PEER) Center \\
        University of California, Berkeley \\
        Berkeley, California, U.S.A \\
        yuanxj23@mails.tsinghua.edu.cn; yuan.xinjie@berkeley.edu
    }

\and
    \IEEEauthorblockN{Khalid M. Mosalam$^*$}
    \IEEEauthorblockA{Pacific Earthquake Engineering Research (PEER) Center \\
        University of California, Berkeley \& \\
        Department of Civil and Environmental Engineering \\
        University of California, Berkeley \\
        Berkeley, California, U.S.A\\
        mosalam@berkeley.edu
    }
}

\maketitle

\let\thefootnote\relax
\footnotetext{Professor Khalid M. Mosalam is the corresponding author. 

This paper has been accepted by and is to appear in \textit{Computer-Aided Civil and Infrastructure Engineering}.}

\begin{abstract}
Fire safety is crucial for ensuring the stability of building structures, yet evaluating whether a structure meets fire safety requirement is challenging. Fires can originate at any point within a structure, and simulating every potential fire scenario is both expensive and time-consuming. To address this challenge, we propose the concept of the Most Fire-Sensitive Point (MFSP) and an efficient machine learning framework for its identification. The MFSP is defined as the location at which a fire, if initiated, would cause the most severe detrimental impact on the building’s stability, effectively representing the worst-case fire scenario. In our framework, a Graph Neural Network (GNN) serves as an efficient and differentiable agent for conventional Finite Element Analysis (FEA) simulators by predicting the Maximum Interstory Drift Ratio (MIDR) under fire, which then guides the training and evaluation of the MFSP predictor. Additionally, we enhance our framework with a novel edge update mechanism and a transfer learning-based training scheme. Evaluations on a large-scale simulation dataset demonstrate the good performance of the proposed framework in identifying the MFSP, offering a transformative tool for optimizing fire safety assessments in structural design. All developed datasets and codes are open-sourced online.

\end{abstract}


\section{Introduction}
\label{introduction}
With the rapid urbanization and increasing complexity of building structures, fire safety has become a critical global issue. Fires pose significant challenges to structural integrity and stability of buildings, making it essential to incorporate fire safety measures during the design phase to enhance resilience and prevent catastrophic failures. The devastating impact of recent wildfires in California highlights the urgency of improving fire risk assessment and mitigation measures in the built environment \cite{cal_fire2023}.

In designing a building structure, there are many requirements to be met, e.g., requirements related to gravity load stability \cite{ASCE7}, earthquake resistance, wind effects, and fire safety \cite{EN1991_1_2}. Before a final check with multiple load combinations, e.g., combination of gravity, earthquake and fire loads, each individual requirement should be satisfied in a pre-check phase. A standard fire safety requirement mandates that a building must withstand collapse for at least one hour under fire conditions at any specific point within the structure. Current methods to ensure this requirement involve computational simulations, e.g., using Finite Element Analysis (FEA) tools, where fire scenarios are simulated at all potential locations within the building structure. This exhaustive approach is computationally expensive and time-consuming, greatly hindering the iterative process in the preliminary design of structures. 

To address this inefficiency, we first propose the concept of the Most Fire-Sensitive Point (MFSP) --the location where fire has the greatest impact on a structure's lateral stability. The MFSP corresponds to the worst-case scenario when a fire occurs. By identifying the MFSP, designers may only need to simulate its corresponding fire scenario to evaluate the structural compliance in the pre-check phase, significantly reducing computational costs and effectively streamlining the design process. However, identifying the MFSP is not a trivial task --conventionally, brute-force methods with computational simulations are still required. Fortunately, recent Machine Learning (ML) advancements provide new opportunities to efficiently identify the MFSP as conducted in this study. Regarding the fire scenarios, \cite{nan_structuralfire_2023} and \cite{wang_graph_2024} employed Recurrent Neural Network (RNN) for real-time displacement prediction, focusing on post-fire collapse warning systems by forecasting future outcomes through early-stage fire features. On the other hand, \cite{chen_tempnet_2024} utilized Graph Neural Network (GNN) for post-fire concrete temperature field mapping with material-level assessment. Despite these works, using ML techniques to assess the overall structural lateral stability in the design stage without real-time information of fires remains a major challenge. 

Moreover, even with a Neural Network (NN) predictor, it is difficult to directly apply traditional supervised learning for MFSP prediction. Traditional supervised learning requires large amount of labeled data, i.e., structures with ground truth MFSP labels. However, obtaining such labeled data is impractical due to the required brute-force simulations, which are both labor-intensive and computationally prohibitive. To overcome this challenge, we propose a framework with two GNN-based predictors: the Maximum Interstory Drift Ratio (MIDR) predictor as the differentiable agent of fire simulations to assess the overall structural lateral stability, and the MFSP predictor as the ``argmaxer'' of the MIDR predictor to predict the MFSP. 

The workflow begins with FEA simulations for a limited set of fire scenarios, which are used to train the MIDR predictor. 
In traditional FEA simulations, the complex operations due to nonlinear material behavior, iterative solvers, and discrete time stepping make traditional FEA tools non-differentiable as well as computationally expensive. In contrast, our differentiable agent uses a GNN with differentiable operations (e.g., a ReLU-activated Multi-Layer Perceptron (MLP)), ensuring computations of the gradients for end-to-end training. Compared to traditional physics-based simulations, the GNN-based MIDR predictor offers two major advantages: (1) the {\em{differentiability}} enables the agent to work as an indicator to directly guide the training of the MFSP predictor to be an ``argmaxer'' of the agent, and (2) the {\em{computational efficiency}} enables it to generate large synthetic data to assist in the training, allowing it to efficiently identify the MFSP without requiring extensive real-world data.

This study focuses on framed steel building structures, which are widely used in modern construction, including commercial facilities. However, the proposed research method is not limited to steel structures and can be extended to other types of buildings with similar structural configurations. The primary contributions of this research are as follows:
\begin{itemize}
    \item {\bf{Concept of MFSP \& workflow for its prediction}}: We first propose the concept of MFSP, which can greatly enhance the structural design process by reducing time and computational cost to check whether fire requirements are satisfied. We also propose a novel  workflow to predict the MFSP for a building structure, including a differentiable neural agent for fire dynamics and FEA simulation and an MFSP predictor trained with the aid of the agent.
    \item {\bf{Transfer Learning (TL) \& Edge Update (EU) in GNNs}}: We propose a GNN-based framework to model building structures and incorporate TL for efficient training. An Edge Update (EU) mechanism is introduced to account for the changes in the material properties under fire scenarios.
    \item {\bf{Dataset generation \& verification of proposed workflow}}: We generate a large-scale dataset of building structures, encompassing their geometry, material properties, and nodal displacements under gravity loads. After filtering out unreasonable structures in the whole dataset, 16,050 structures form the {\em{unlabeled}} dataset, and another 1,573 structures are simulated for 30 fire scenarios, forming the {\em{labeled}} dataset, which contains the nodal displacements for each of the $1,573 \times 30 = 47,190$ fire cases, obtained using a FEA tool, namely, $\chi$ara \cite{perez2024xara} (previously named OpenSeesRT \cite{perez2024openseesrt}). Simulation results based on the generated dataset verify the efficiency and accuracy of the proposed workflow. Ablation experiment also demonstrates the effectiveness of the proposed TL and EU mechanisms.
    \item {\bf{Open-source resources}}: We provide the dataset and implementation details via a public GitHub repository: \url{https://github.com/STAIRlab/MFSP_Prediction}.
\end{itemize}

After this introduction, the paper is organized as follows: \secref{sec:literature} reviews relevant literature on GNN applications, agents, and fire analysis of buildings. \secref{sec:system_overview} provides an overview of the proposed system architecture. Sections \ref{sec:mdrp}, \ref{sec:mfspp}, and \ref{sec:dataset} detail the MIDR predictor, the MSFP predictor, and the dataset generation, respectively. Implementation details and evaluation results are presented in \secref{sec:evaluation}. We discuss the limitations and proposed future work in \secref{sec:discussions}. Finally, \secref{sec:conclusion_future} is a concise summary of the main conclusions of the study. Through this research, we aim to enhance fire risk assessment accuracy and efficiency, providing a robust tool to support the design of safer, more resilient buildings.

\section{Literature Review}
\label{sec:literature}
Understanding the interaction between fire dynamics and building structures is fundamental to designing safer, more resilient building systems. Resilience, in this context, encompasses robustness, adaptability, recoverability, and redundancy, ensuring systems can withstand, adapt to, and recover from adverse events. Recent advances in FEA, ML, and physics-informed simulations have provided powerful tools for analyzing and predicting the structural performance under fire conditions. This section reviews key contributions in these domains, identifies gaps, and motivates the integration of differentiable agents with thermal simulators for enhanced fire analysis in buildings.

\subsection{GNN Applications and Agents}
Recently, GNN has shown successful applications in many areas in civil engineering. For example, \cite{huang_dynamic_2024} employed GNN for predicting network-wide metro passenger flow with dynamic propagation graphs, and \cite{sheng_egoplanningguided_2024} introduced multi-graph convolutional network designed to predict the future trajectories of traffic agents near an autonomous vehicle by incorporating various interaction perspectives and EGO-planning information, where EGO stands for {\bf{E}}uclidean Signed Distance Field-free {\bf{G}}radient-based l{\bf{O}}cal. Moreover, \cite{gao_urban_2024} utilized physics-informed GNN-assisted auto-encoder to reconstruct high-resolution urban wind fields based on sparse sensor data. \cite{jia_graph_2023a} reviewed the application of GNNs in construction. These successful applications demonstrate the strong ability of GNN to capture spatial relationships and the potential for applying GNN in structural analysis. In predicting structural responses, \cite{chou_structgnn_2024} applied GNN to a static problem to predict responses such as nodal displacements and shear forces, showing very high accuracy. GNN was further combined with RNN in \cite{jia_temporal_2024} to predict building thermal load in general scenarios, focusing on capturing spatial interactions between multiple thermal zones and temporal dependencies. 

Despite these successful applications, using GNN to predict the structural response in fire scenarios has not been well studied. \cite{wang_graph_2024} proposed a GNN-RNN hybrid model for real-time displacement prediction of steel frames under fire, yet the work focused on post-fire collapse warning systems by forecasting future outcomes through early-stage fire features, rather than evaluating the stability at the design phase. Meanwhile, \cite{chen_tempnet_2024} developed a GNN-based model for post-fire concrete temperature field mapping, but their approach remained limited to material-level assessment and did not address the response at the structural level. Moreover, existing works merely use GNN for inference, i.e., prediction, neglecting that it could also work as an \emph{agent} for other purposes. The concept of agent has been widely used in civil engineering applications. For example \cite{soto_multiagent_2017} proposed a multi-agent replicator controller for vibration control of smart structures, and \cite{yao_multiagent_2024} introduced a multi-agent Reinforcement Learning (RL) model for optimizing maintenance decisions in interdependent highway pavement networks. However, these works mainly treat agents as the final objective, i.e., the well-trained agents are directly used to generate the desired decisions. We distinguish the presented study from these works by the role of the agent in the framework, where herein the agent works as a {\em{surrogate model}} and is an {\em{intermediate module}} in the proposed framework. The agent's differentiability and high efficiency are fully utilized in the presented work.

\subsection{Thermal Simulation}
Fire event simulation encompasses several critical aspects, including heat transfer modeling, fire spread dynamics, and the impact of thermal loads on structural elements.
The interplay between thermal loads and structural elements during fire events has been extensively studied. Early work in \cite{anderberg1988modelling} introduced a method for simulating heat transfer in steel structures. This approach was later enhanced to incorporate transient thermal responses and load redistribution effects \cite{kodur2010response}, which remain vital for predicting collapse mechanisms under high-temperature conditions \cite{franssen2002fire}.

Beyond heat transfer analysis, fire simulation also involves assessing the deterioration of structural resistance. This includes temperature-induced material degradation, progressive weakening of load-bearing components, and failure mechanisms under prolonged fire exposure. Notably, steel structures experience reductions in Young’s modulus and yield strength at elevated temperatures, which affect their ability to sustain applied loads. Thermal expansion and differential heating further contribute to internal stress redistribution, potentially leading to local or global instability. Although phenomena such as buckling of slender members, loss of composite action in beam-column connections, and creep effects under sustained high temperatures are integral considerations in advanced fire analysis models, they are beyond the scope of the present paper and should be considered in future studies. 

Recently, \cite{wang_optimizationimproved_2024} developed a computational framework to enhance the simulation of welding residual stresses using ML and optimization techniques, validated through experimental measurements. \cite{wan_thermal_2024} studied the thermal contraction coordination behavior between an unbound aggregate layer and an asphalt mixture overlay using a finite difference and discrete element coupling method. Most past works focused on local behavior of system components under thermal loads instead of the overall structural stability.

Computational Fluid Dynamics (CFD)-based methods, such as Fire Dynamics Simulator (FDS) \cite{mcgrattan_fire_2000}, provide high-fidelity modeling of temperature evolution, flame spreading, and gas-phase combustion. However, the computational cost remains a significant limitation, restricting large-scale structural assessments. FDS requires not only a large computing time for the simulation, but also significant modeling time for the building structures. This limits the use of FDS to case studies with necessary input of building structural details, as well as information on combustible materials inside the building. For example, \cite{manea_fds_2022} employed FDS modeling to investigate two ignition scenarios in 2014 Romanian restaurant fire, determining the most likely cause of rapid fire spread and lethal Carbon Monoxide exposure that resulted in four fatalities.

Real-world fire loads often display substantial spatial variability across different rooms in a building \cite{dundar_fire_2023}, resulting in scenario-specific temperature fields with limited generalizability. Furthermore, even within identical scenarios, variations in fire modeling methodologies can produce distinctly different temperature fields, as demonstrated by \cite{zhang_temperature_2020} using a full-scale experiment to develop a new temperature field prediction model for large space fires, and by \cite{du_new_2012} through creating a new temperature-time curve to better represent the non-uniform temperature distribution in large space fires compared to standard curves. Moreover, studies on bridge fires \cite{he_study_2024} demonstrated that environmental factors, such as wind speeds, can significantly influence the temperature distributions. These challenges emphasize the need for efficient and adaptive methods to generate fire temperature data, where less important factors might be neglected. 

While existing studies have extensively explored post-fire structural behavior, integrating material deterioration mechanisms into a ML-based predictive framework remains an open challenge. The proposed approach in this study aims to bridge this gap by efficiently approximating fire-induced structural responses without relying on computationally intensive physics-based simulations. To overcome the above difficulties, we propose a {\em{rule-based}} temperature generation method, which is efficient and scalable for serving large-scale dataset simulations. With a generated temperature field in a structure, FEA tools,  e.g., \cite{yu_objectoriented_1993}, can be used to obtain the structural response under thermal loads. For steel structures, previous studies, e.g., \cite{nan_structuralfire_2023}, have demonstrated that they rapidly equilibrate with the surrounding gas temperatures due to efficient heat exchange. Consequently, gas temperatures can be directly used as input for FEA. This inspired us to generate a large dataset using FEA for thermal analysis in fire scenarios to form the foundation of the NN training and testing. 
\section{System Overview}
\label{sec:system_overview}
\begin{figure*} [h!]
    \centering
    \includegraphics[width=0.8\textwidth]{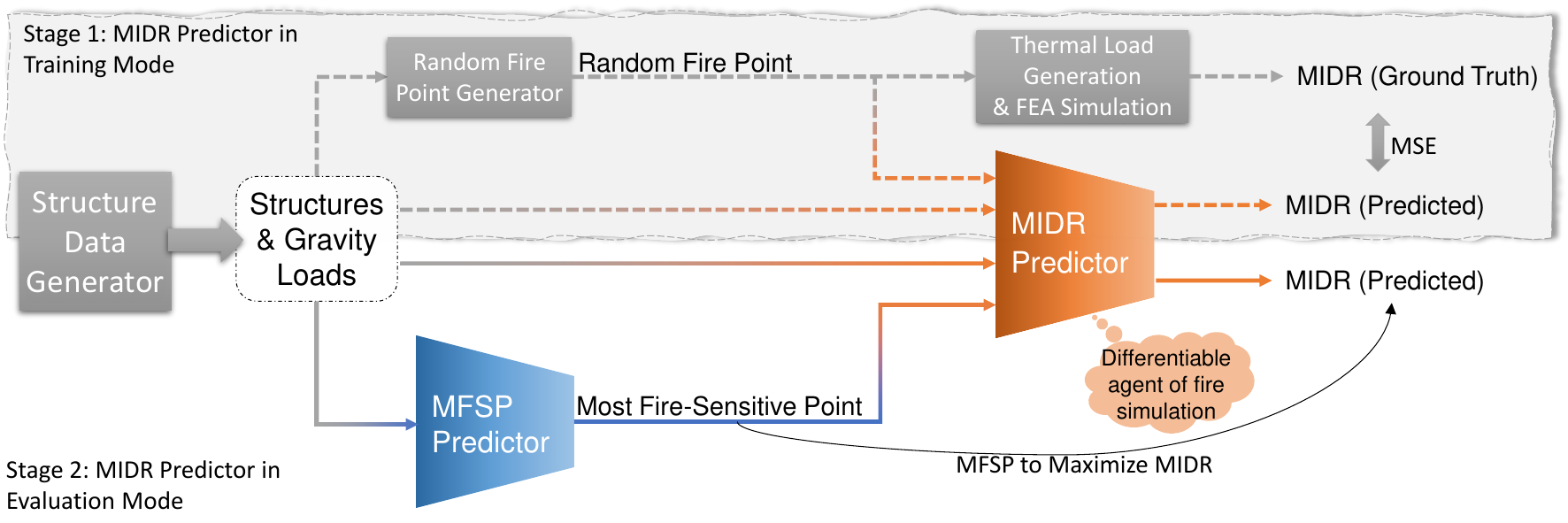}
    \caption{Proposed framework for predicting the MFSP in building structures. Trapezoids  and gray rectangles represent the NN predictor and processing, respectively. Dashed and solid lines indicate that MIDR predictor is in the respective training mode to determine its parameters and evaluation mode with parameters fixed, acting as a differentiable agent.}
    \label{fig:system_overview}
\end{figure*}

The primary goal of this work is to accurately predict the MFSP in building structures subjected to various fire scenarios by utilizing the MIDR as a metric for the overall lateral stability. To achieve this, we propose an integrated framework that combines GNNs and FEA. The system architecture is illustrated in \figref{fig:system_overview} and comprises two key components: the MIDR predictor and the MFSP predictor. There are two stages corresponding to the MIDR predictor's different modes. In stage 1, we use data with ground truth MIDR values obtained from FEA simulations to train the MIDR predictor. Then, in stage 2, with the well-trained MIDR predictor working as a differentiable agent of fire simulation, we train the MFSP predictor to determine the point that maximizes the MIDR. This framework seamlessly integrates physics-based simulations, GNN-driven predictive modeling, and data-driven techniques, enabling efficient and accurate MFSP prediction to provide a powerful tool for proactive fire safety analysis and risk mitigation in building structures.

\subsection{Structural Stability Metrics} 
The Interstory Drift Ratio (IDR) of each node serves as a critical parameter for evaluating structural lateral stability and deformation under external forces. IDR quantifies the relative displacement between two consecutive floors (interstory displacement) as a percentage of the floor height. Mathematically, the IDR for a given node $i$ is defined as follows:
\begin{equation}
    d_i = \left.\sqrt{\left(\Delta x_i\right)^2 + \left(\Delta y_i\right)^2}\right/ H \times 100 \%,
\end{equation}
where the numerator represents the relative displacement (in the horizontal plane $X-Y$) of node $i$ with respect to the corresponding node on the floor below, and $H$ denotes the story height between these two floors. Excessively high drift ratios indicate significant structural deformation, potentially leading to major damage or collapse due to lateral instability. The Maximum IDR among all the nodes of a structure, i.e., MIDR, is chosen to be a representative example metric for assessing the overall structural stability performance during fire events. Although MIDR may not capture all possible failure modes, such as local collapses due to midspan softening, local buckling, or loss of vertical elements, we emphasize that the proposed method is not limited to MIDR. The framework can be adapted to other performance indicators of interest, by simply replacing the MIDR with the desired metric.

\textit{Remark:} Selecting MIDR as the primary metric for assessing the structural integrity under fire conditions is motivated by its effectiveness in quantifying global deformation patterns. In fire-induced scenarios, thermal expansion, stiffness degradation, and gravity-induced deformations contribute to structural instability, which can be captured through relative floor displacements. MIDR provides a direct and interpretable measure of structural vulnerability by identifying floors experiencing excessive lateral deformations that may lead to global instability or loss of vertical load-carrying capacity. While MIDR is commonly used for seismic and wind-induced responses, its application to fire scenarios is justified as a fire-driven thermal effect also leads to large-scale deformation patterns, particularly in multi-story steel structures. Importantly, MIDR serves as an effective proxy for overall building stability in computational frameworks where parameterizing every potential failure mode (e.g., local buckling, connection failure, progressive collapse) is not feasible. However, fire-induced failure mechanisms extend beyond interstory drift, and localized effects are not explicitly captured by MIDR. These mechanisms typically develop locally and may not always translate into immediate global structural instability. While our current framework focuses on identifying the MFSP based on a worst-case drift metric, future work could integrate alternative failure criteria to further refine the fire vulnerability predictions.

Note that ``M'' in MIDR and MFSP represents different concepts. In the case of MIDR, for a given structure and fire source point, the IDR is computed at each node, and the MIDR is defined as the maximum IDR among all nodes. In contrast, MFSP refers to the fire source location that results in the highest MIDR across all possible fire source points within the structure. In summary, an MIDR is associated with a specific structure and fire source point pair, whereas an MFSP characterizes an entire structure by identifying the most critical fire source location.

\subsection{MIDR \& MFSP Predictors}

The MIDR predictor is a GNN-based model designed to estimate the MIDR of a building under a given fire scenario. The inputs to this model include:
\begin{itemize}
    \item {\bf{Structural configuration}}: Building geometry, material property, and gravity loads. 
    \item {\bf{Fire location}}: The specific point where the fire is initiated within the building.
\end{itemize}
A GNN processes this input to represent the structural configuration and fire location as a graph. The MIDR predictor is trained on labeled data generated using $\chi$ara \cite{perez2024xara}, a robust open-source FEA framework based on OpenSees \cite{jiang_opensees_fire_2015}. These labels represent detailed structural responses under various fire conditions. Once trained, the MIDR predictor functions as a {\em{differentiable agent}}, offering computationally efficient MIDR estimates. Its capabilities include:
\begin{itemize}
    \item {\bf{Annotating datasets}}: Assigning  MIDR values to support subsequent analyses.
    \item {\bf{Integrating with NNs}}: Reducing the computational cost typically associated with simulation-based methods.
\end{itemize}

The MFSP predictor acts as an ``argmaxer module'' for the MIDR predictor, identifing the fire location that results in the highest MIDR. This location corresponds to the point of the greatest structural vulnerability. By leveraging the structural graph as input and utilizing the MIDR predictor's outputs, the MFSP predictor efficiently pinpoints the critical fire location.

\subsection{Data Generation and Training Pipeline}
To ensure robustness and generalizability, we introduce a comprehensive data generator pipeline:
\begin{enumerate}
    \item {\bf{Structure data generator}}: This component creates synthetic datasets for diverse building configurations, including geometry, material, and gravity loads.
    \item {\bf{FEA simulations}}: With the high-fidelity FEA simulation software, $\chi$ara, the gravity simulation is first conducted to confirm the rationality of the synthetic dataset. Further, a subset of the generated configurations undergoes fire scenario simulations using $\chi$ara based on a rule-based thermal load generation method. These simulations produce {\em{labeled data}} detailing  the structural responses to various fire locations, forming training and testing sets for the MIDR predictor.
    \item {\bf{Unlabeled data utilization}}: The remaining configurations, without MIDR labels from the FEA simulations, are also used to train and test the MFSP predictor, leveraging the MIDR predictor as a computationally efficient, yet accurate, {\em{surrogate}} model. Although the structural configurations with unlabeled data do not undergo FEA simulations, they can be rapidly and efficiently  \emph{pseudo labeled} using this surrogate model.
\end{enumerate}

\section{MIDR Predictor}
\label{sec:mdrp}
The MIDR predictor functions as a differentiable agent for the FEA simulators. Leveraging GNNs for structural representation, the predictor integrates Transfer Learning (TL) to optimize data utilization while focusing on the MIDR prediction. Additionally, an innovative Edge Update (EU) module, to update edge attributes of the GNNs, is introduced to capture changes in the finite element attributes during fire scenarios.

\subsection{GNNs}
\label{subsec:gnn}

\subsubsection{Overview}
\begin{figure*}[h!]
    \centering
    \includegraphics[width=1\linewidth]{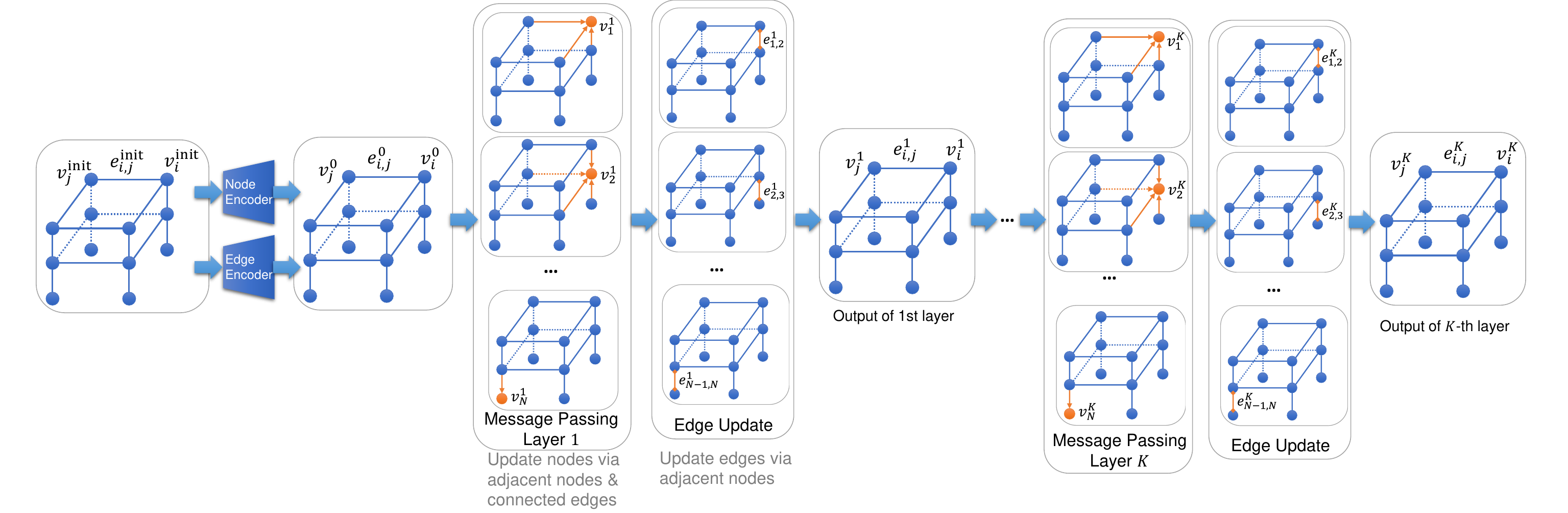}
    \caption{Illustration of a $K$-layer GNN architecture, demonstrating information flow within the network for a graph with $N$ nodes.}
    \label{fig:gnn_illustration}
\end{figure*}

Building structures and graphs share an intrinsic similarity: structural nodes correspond to graph nodes, and structural elements (e.g., beams and columns) map naturally to graph edges. This analogy forms the basis for utilizing GNNs in structural data processing. A typical GNN consists of multiple stacked layers, where the output of one layer serves as the input for the next. Each  layer executes three core operations --message passing, aggregation \& update-- expressed as follows:
\begin{equation}
    \label{eq:gnn_overview}
    \vec{v}_i^k = \gamma^k \left( \vec{v}_i^{k-1}, \bigoplus_{j \in \mathcal{N}(i)} \phi^k \left( \vec{v}_i^{k-1}, \vec{v}_j^{k-1}, \vec{e}_{j,i} \right) \right),
\end{equation}
where $\gamma^k(\cdot)$ and $\phi^k(\cdot)$ represent the update and message functions, respectively, $\bigoplus$ is the aggregation operation, e.g., sum, mean, or max, $\mathcal{N}(i)$ denotes the set of neighboring nodes of node $i$, and $\vec{v}_i^k$ and $\vec{e}_{j,i}$ are the attributes of node $i$ at the $k$-th layer and attributes of the edge from node $j$ to node $i$, respectively. Here, both $\gamma^k(\cdot)$ and $\phi^k(\cdot)$ are MLPs, i.e., fully connected NNs. The operations $\gamma(\cdot) ,\phi(\cdot) \, \& \, \bigoplus$ are not related to certain nodes or edges, indicating that the GNN architecture inherently supports variable-sized graphs through parameter-sharing mechanisms. Each graph is processed independently, with identical NN operations applied to all nodes and edges regardless of their number. \figref{fig:gnn_illustration} is a schematic of a GNN with $K$ layers, illustrating the role of node and edge encoders, as well as the EU mechanism. The shown GNN incorporates initial node and edge encoders to transform the input features into higher-dimensional spaces and utilizes an EU mechanism to dynamically update the edge attributes during each layer to capture changes of the finite elements during fire scenarios.

\subsubsection{Input attributes}
\label{subsubsec:input_attributes}
In our framework, gravity and thermal loads are applied to the beams and columns. These elements dictate the information encapsulated in the nodes and edges of the graph representation.

\paragraph{Node attributes} 
The coordinates of the $i$-th node in 3 Dimensions (3D) $(x_i, y_i, z_i)$, essential to represent the spatial configuration of the structure, are included as node attributes. Additionally, its floor level $h_i$, a critical factor to define the fire scenarios, is included as another node attribute. For ease of description, we call the four-tuple $\left(x_i, y_i, z_i, h_i\right)$ extended coordinates of node $i$. To account for the fire location, the extended coordinates of the fire source $\left(x_{\subfire}, y_{\subfire}, z_{\subfire}, h_{\subfire}\right)$, containing the 3D coordinates and the corresponding floor level of the fire, are incorporated into the node attributes. Note that $h_i$ and $h_{\subfire}$ are integers.
To enhance the model's ability to capture fire-related information, we compute the differences between the node's and fire source's extended coordinates $\left(x_i - x_{\subfire}, y_i - y_{\subfire}, z_i - z_{\subfire}, h_i - h_{\subfire}\right)$.
Finally, the Euclidean distance between the $i$-th node and the fire location is calculated as $\sqrt{(x_i-x_{\subfire})^2 + (y_i-y_{\subfire})^2 + (z_i - z_{\subfire})^2}$. These features collectively form the initial attributes of each node $i$, denoted as $\vec{v}_{i}^{\supinit}$, comprising a total of 13 features, i.e., $4$ for the extended coordinates of node $i$, $4$ for the extended coordinates of the fire source, $4$ for the differences between these two sets of the extended coordinates, and $1$ for the Euclidean distance. Note that since we know that fire propagation is strongly correlated with distance and building height, incorporating these features as inputs to the NN is a reasonable approach, which does not complicate the training process, but improves its reliability instead.

\paragraph{Edge attributes} The edges between nodes represent the structural elements (beams and columns) and encapsulate the following attributes:
\begin{itemize}
    \item {\bf{Material properties}}: Young's modulus \& yield strength at ambient temperature, and the strain-hardening ratio, counting up to 3 dimensions.
    \item {\bf{Geometric properties}}: Length, floor level, and element orientation. Note that the orientation is encoded using a {\em{one-hot representation}} for alignment along the $x$-axis, $y$-axis, or $z$-axis. Therefore, the geometric properties count up to 5 dimensions. In this study and for simplicity, we use square sections for the beams and columns, requiring no additional attributes for the orientation of these sections.
    \item {\bf{Gravity loads}}: The magnitude of the applied load on the element counts as the 9-th dimension.
\end{itemize}
The initial attributes of each edge $(i,j)$, denoted as $\vec{e}_{i,j}^{\supinit}$, are represented by 9-dimensional vector combining the above three sets of attributes.

\paragraph{Feature encoding} The raw node and edge attributes are not directly input into the GNN. Instead,  encoders are employed to map the initial attributes into higher-dimensional spaces, which improve the model's ability for feature representation as follows: 
\begin{itemize}
    \item {\bf{Node encoder}}: Extends the 13-dimensional node attributes $\vec{v}_{i}^{\supinit}$ to $\vec{v}_{i}^0$ with a higher dimensionality, e.g., 64 dimensions. The superscript $0$ implies input of the first GNN layer.
    \item {\bf{Edge encoder}}: Maps the 9-dimensional edge attributes $\vec{e}_{i,j}^{\supinit}$ to $\vec{e}_{i,j}$ with a higher-dimensional representation, e.g., 32 dimensions. 
\end{itemize}
After extending the dimensions of node and edge features using encoders, their dimensions remain consistent across different layers of the GNN. This simplifies the architecture of the NN design, where different layers have similar structures, all with the same input and output dimensions. This characteristic of unchanged dimensions allows for the ease of extracting edge or node features from any layer of the GNN and performing the same processing for the MIDR prediction, as detailed in \secref{subsubsec:handle_structural_data}. If we do not use encoders to expand the dimensions and accordingly maintain consistent dimensions across different layers of the GNN, e.g., keeping node attributes at 13 dimensions, the model capacity decreases and the performance of the predictor declines.

\subsubsection{Handling structural data}
\label{subsubsec:handle_structural_data}
A significant challenge in GNNs is the {\em{over-smoothing}}, where during each layer's update process, a node's attributes aggregate information from its neighboring nodes. Consequently, increasing the number of layers causes node attributes to homogenize, losing their specificity and  negatively impacting the model's performance. This issue is particularly pronounced in building structures with varying story counts and number of nodes, such as 2-story vs. 7-story buildings. If both are passed through a 7-layer GNN, the node attributes of the 2-story building would experience severe over-smoothing. 
In simple terms, when there are too many message-passing layers, the characteristics of all nodes tend to be consistent. However, too few layers lead to the problem of incomplete perception, i.e., nodes cannot obtain information from distant nodes.
To mitigate this problem, we adopt the method from \cite{chou_structgnn_2024}, restricting the number of GNN layers for each graph, i.e., building structure herein, to the building's number of stories. 
In this way, the number of layers is properly set, neither too many nor too few, ensuring that critical information about fire propagation is retained in the node attributes without over-smoothing.

Beams and columns exhibit distinct loading  characteristics that influence how they are represented in graph-based model of a structure. Beams primarily bear gravity loads that act perpendicular to their span direction, which aligns with their structural function in predominantly resisting bending moments and shear forces. Under thermal expansion due to fire, these loads affect both end nodes of the beam. As a result, when representing beams in a graph model, it is intuitive to treat them as {\em{undirected edges}}, reflecting their bidirectional influence on the connected nodes. Columns, on the other hand, experience gravity loads along their axial direction, directed downward, which is consistent with their role in mainly supporting vertical loads in addition to bending moments and shear forces due to their frame action with the floor beams. When a column deforms under loads (due to thermal expansion, gravity, or lateral loads), the displacements at any point in the column influence the displacements above it, which continues sequentially upward along the column length. However, in the mathematical formulation of the FEA, the computations involve solving the global system of equations that inherently account for all inter-dependencies between elements and boundary conditions simultaneously, not sequentially. Accordingly, representing columns as {\em{directed edges}} might initially seem appropriate, as it allows separate transmission of force and displacement information. However, in GNNs, undirected graphs are commonly implemented by duplicating directed edges and reversing their directions to ensure bidirectional information flow. Given this practical implementation in GNNs, both beams and columns are treated as {\bf{undirected edges}} in our graph model. This approach simplifies the representation while preserving the necessary information flow for structural analysis.

\subsubsection{Edge update mechanism}
\label{subsubsec:edge_update}
Typically, GNNs often do not update edge attributes, $\vec{e}_{i,j}$, during the iterative update process in each layer. However, in the context of fire scenarios, the attributes of beams and columns (e.g., material properties and geometrical characteristics) evolve with temperature changes, directly affecting the MIDR computations. To address this need, we introduce an EU module that dynamically updates edge attributes during each layer's computations. After updating node attributes via message passing in each layer of the GNN, the EU module utilizes the attributes of the two adjacent nodes and the edge itself to update the edge attributes. This ensures that the edge features reflect the ongoing structural changes due to fire conditions. 

With the EU module, the standard GNN operation in Equation \eqref{eq:gnn_overview} is modified to include EUs. The updated node attributes are expressed as follows:
\begin{equation}
    \label{eq:gnn_with_edge_update}
    \vec{v}_i^k = \gamma^k \left( \vec{v}_i^{k-1}, \bigoplus_{j \in \mathcal{N}(i)} \phi^k \left( \vec{v}_i^{k-1}, \vec{v}_j^{k-1}, \vec{e}_{j,i}^{k-1} \right) \right).
\end{equation}
The EU module updates the edge attributes as follows:
\begin{equation}
    \label{eq:edge_update}
    \vec{e}_{j,i}^k = \zeta \left( \vec{v}_i^{k}, \vec{v}_j^{k} \right) + \vec{e}_{j,i}^{k-1},
\end{equation}
where $\zeta(\cdot)$ is the update function for edge attributes, implemented as a single-hidden-layer MLP. The inclusion of the EU module enables:
\begin{enumerate}
    \item {\bf{Dynamic adaptation}} capturing temperature-induced changes in beam and column properties during each GNN layer's computations.
    \item {\bf{Enhanced modeling}} simultaneously updating node and edge attributes results in a more accurate representation of the structural state.
    \item {\bf{Improved performance}} providing the GNN with the ability to better model the structural transformations under fire scenarios leads to more precise MIDR predictions.
\end{enumerate}    
By iteratively updating edge attributes, the EU module effectively integrates the evolving structural properties into the GNN, enhancing its predictive capabilities in fire scenarios.

\subsection{TL-based Network Training}
\label{subsec:transfer_learning}
The proposed GNN framework only computes an embedding vector for each node. Therefore, to obtain the predicted MIDR, additional operations (layers) are needed. Two straightforward methods to aggregate the node-level results to a graph-level prediction can be applied. These two baseline methods are referred to as Strawman 1 \& 2 and they only differ in where and what to aggregate, as discussed below:
\begin{itemize}
    \item {\textbf{Strawman 1}} aggregates the node embeddings in a graph into a single graph embedding, using a {\em{pooling operation}}, which predicts the MIDR with an additional MLP.
    \item {\textbf{Strawman 2}} predicts the IDR of each node using node embeddings and an additional MLP. Then, it aggregates the node-level IDR predictions by selecting their MIDR.
\end{itemize}

\begin{figure}[h!]
    \centering
    \includegraphics[width=1\linewidth]{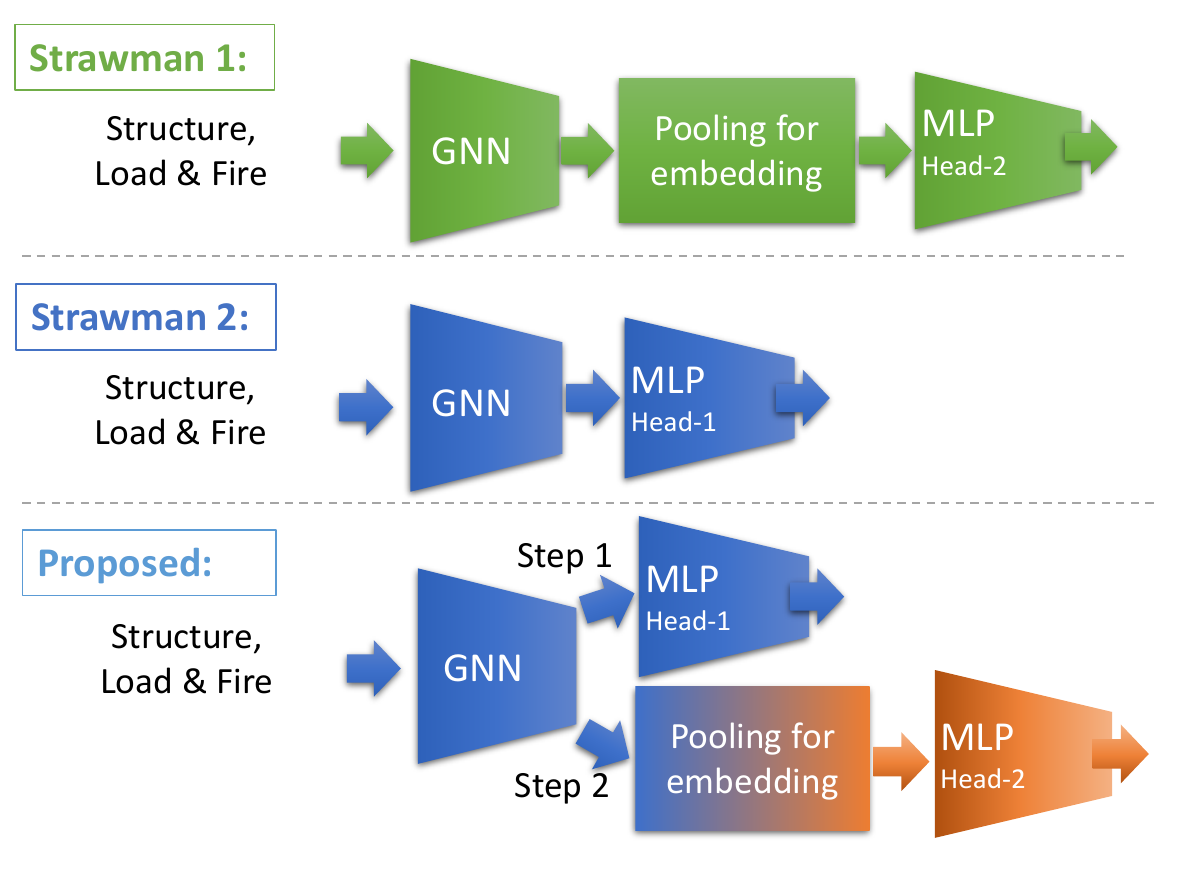}
    \caption{Different training strategies (head-1 \& head-2 are MLPs for different tasks, e.g., predicting IDR of each node in a structure and MIDR of the structure).}
    \label{fig:transfer_learning}
\end{figure}

While Strawman 1 focuses solely on predicting the MIDR, it fails to leverage the IDR labels data available for individual nodes. Consequently, this approach risks overfitting when the dataset is small. On the other hand, Strawman 2 uses all node-level IDR labels, but the inclusion of non-MIDR nodes during training dilutes the model's focus on predicting the MIDR. To overcome these limitations, we propose a TL-based training method that integrates the strengths of both approaches. By framing the problem as two related tasks --predicting node-level IDRs and structure-level MIDR-- we adopt a multi-task TL approach with a two-step training procedure, \figref{fig:transfer_learning}. 

In the first step, the GNN is trained to predict the IDR of individual nodes, using an MLP referred to as {\bf{task head-1}} as in Strawman 2, \figref{fig:transfer_learning}. During this step, the GNN learns to extract global structural features. The loss function for this step is defined as follows:
\begin{equation}
    \label{eq:mdr_loss_task1}
    \begin{aligned}
        & L_{\text{task-1}}\left(\Theta_{\text{GNN}}, \Theta_{\text{head-1}}\right)\\
        & \qquad=\frac{1}{\sum_{m=1}^{M} N_{m} I_{m}}\sum_{m=1}^{M} \sum_{n=1}^{N_{m}} \sum_{i=1}^{I_{m}} \left( d_{m,n,i}^{\, \gt} - \widehat{d}_{m,n,i} \right)^2,
    \end{aligned}
\end{equation}  
where $d_{m,n,i}^{\, \gt}$ is the ground truth IDR for node $i \in \left(1,2,\dots,I_{m}\right)$ in structure $m \in (1,2,\dots,M)$ computed via $\chi$ara under fire scenario $n \in \left(1,2,\dots,N_{m}\right)$, $\widehat{d}_{m,n,i}$ is the predicted IDR from the model, $\Theta_{\text{GNN}}$ is the set of parameters of the GNN, and $\Theta_{\text{head-1}}$ is the set of parameters of task head-1. The optimization objective for this step is as follows:  
\begin{equation}
    \label{eq:mdr_task1}
    \min_{\Theta_{\text{GNN}}, \Theta_{\text{head-1}}} L_{\text{task-1}}\left(\Theta_{\text{GNN}}, \Theta_{\text{head-1}}\right).
\end{equation}  

In the second step, the trained GNN parameters from the first step are reused as a {\em{feature extractor}}. The MLP for node-level IDR prediction (task head-1) is replaced with a new MLP, referred to as {\bf{task head-2}} as in Strawman 1, \figref{fig:transfer_learning}, to predict the structure MIDR. During this step, task head-2 is first trained independently, followed by {\em{fine-tuning}} the entire GNN. The MIDR for structure $m$ under fire scenario $n$ is defined as:  
\begin{equation}
    d_{m,n,\max} \triangleq \max_{i=1,2,\dots,I_{m}} d_{m,n,i}.
\end{equation}  
The loss function for this step is defined as follows:
\begin{equation}
    \label{eq:mdr_loss_task2}
    \begin{aligned}
        & L_{\text{task-2}}\left(\Theta_{\text{GNN}}, \Theta_{\text{head-2}}\right)\\
        & \qquad=\frac{1}{\sum_{m=1}^{M} N_{m}}\sum_{m=1}^{M} \sum_{n=1}^{N_{m}} \left( d_{m,n,\max}^{\, \gt} - \widehat{d}_{m,n,\max} \right)^2,
    \end{aligned}
\end{equation}  
where $d_{m,n,\max}^{\, \gt}$ is the ground truth MIDR for structure $m$ under fire scenario $n$ computed using $\chi$ara, $\widehat{d}_{m,n,\max}$ is the predicted MIDR from the model, and $\Theta_{\text{head-2}}$ is the set of parameters of task head-2. The optimization objective for this step is as follows:
\begin{equation}
    \label{eq:mdr_task2}
    \min_{\Theta_{\text{GNN}}, \Theta_{\text{head-2}}} L_{\text{task-2}}\left(\Theta_{\text{GNN}}, \Theta_{\text{head-2}}\right).
\end{equation}  

The proposed TL-based methods offers these benefits:
\begin{enumerate}
    \item {\bf{Effective utilization of data}} leverages node-level IDR labels while focusing on MIDR prediction.
    \item {\bf{Improved generalization}} enhances the model's ability to generalize across diverse fire scenarios and structural configurations.
    \item {\bf{Accurate MIDR predictions}} integrates global structural information for precise MIDR predictions, outperforming the individual Strawman methods 1 \& 2.
\end{enumerate}
By combining node-level and structure-level predictions in a two-step training pipeline, the TL-based approach achieves a robust balance between learning granular details and capturing global structural patterns. 

\section{MFSP Predictor}
\label{sec:mfspp}
\begin{figure*}[th!]
    \centering
    \includegraphics[width=1\linewidth]{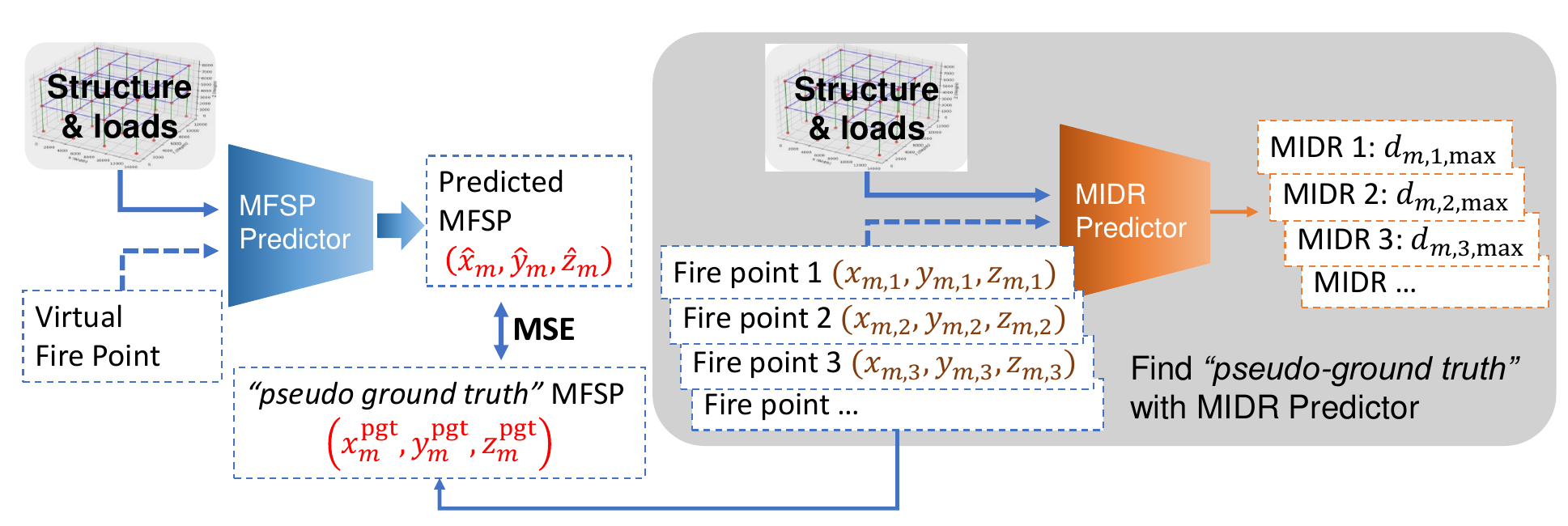}
    \caption{Procedure of pseudo-labeling for building structure $m$.}
    \label{fig:pseudo_label}
\end{figure*}
With the training of the MIDR predictor complete, we have developed a NN-based agent model that effectively serves as a {\em{surrogate}} for the FEA conducted using $\chi$ara. This agent model presents two critical advantages over traditional FEA: (1) computational efficiency and (2) differentiability. These advantages are fully leveraged in training the MFSP predictor, designed to identify the location within the building structure that exhibits the highest vulnerability under fire conditions via the MIDR. Importantly, the training process for the MFSP predictor no longer depends on direct FEA results. Instead, the fixed parameters of the pretrained MIDR predictor are utilized as the computational platform. This allows the MIDR predictor to remain unchanged during the MFSP predictor's training, ensuring consistency and efficiency in the workflow. By relying solely on the outputs of the MIDR predictor, the MFSP predictor benefits from the speed and differentiability of the agent model while eliminating the need for repeated computationally expensive FEA simulations.

\subsection{Pseudo-Labeling and Loss Choice}
\label{subsec:mfspp_pseudo_label}
As illustrated in \figref{fig:system_overview}, the MFSP predictor operates as the ``argmaxer'' of the MIDR predictor. An intuitive choice of the loss function, $L_{\text{MIDR}}$, is to minimize the negative output of the MIDR predictor. This loss leverages the differentiable nature of the MIDR predictor as an agent. It is expressed as:  
\begin{equation}
    L_{\text{MIDR}} = - \frac{1}{\sum_{m}^{M}N_{m}}\sum_{m}^{M}\sum_{n}^{N_m} \widehat{d}_{m,n,\max}.
\end{equation}
This approach conceptually resembles the Actor-Critic framework in Reinforcement Learning (RL) \cite{sutton2018reinforcement} where the MFSP predictor is the ``Actor'', proposing actions in the form of fire points, while the MIDR predictor is the ``Critic'', evaluating the quality of the Actor's actions. However, similar to challenges in RL, this method can lead to poor training outcome or convergence issues. MIDR predictor is a complex function and optimizing its output using gradient descent often results in local optima or unstable training. Additionally, the fixed parameters of the pretrained MIDR predictor limit the ability to introduce randomness for balancing {\em{exploration}} and {\em{exploitation}}, a core mechanism in RL.

To address these issues, we propose an alternative approach inspired by the computational efficiency of the MIDR predictor. Instead of directly using $L_{\text{MIDR}}$, we use the MIDR predictor to generate {\em{pseudo ground truths}} for the MFSP. This process is depicted in \figref{fig:pseudo_label}. Specifically, at the granularity of the rooms, we consider the center point of each room in a building as a potential fire point. The MIDR predictor estimates the MIDR for each fire point, and the one with the highest MIDR is selected as the pseudo ground truth MFSP for the building at hand. This pseudo-labeling approach allows efficient labeling of unlabeled data at the room-level granularity.

For a given building structure $m$, let the pseudo ground truth MFSP coordinates be $\left( x_{m}^{\, \pgt}, y_{m}^{\, \pgt}, z_{m}^{\, \pgt}\right)$, with the corresponding MIDR $\widehat{d}_{m}^{\, \pgt}$. Using this pseudo-labeling, we define the Mean Squared Error (MSE) loss as follows:  
\begin{equation}
    \begin{aligned}
        L_{\text{MSE}} = \frac{1}{M}\sum_{m=1}^{M}&\left( \left( x_{m}^{\, \pgt} - \widehat{x}_{m} \right)^2 + \left( y_{m}^{\, \pgt} - \widehat{y}_{m} \right)^2 \right. \\  
        & \left.+ \left( z_{m}^{\, \pgt} - \widehat{z}_{m} \right)^2 \right),
    \end{aligned}
    \label{eq:MSE}
\end{equation}
where $\widehat{x}_{m}$, $\widehat{y}_{m}$, and $\widehat{z}_{m}$ are the MFSP predictor's outputs for structure $m$. Since the pseudo-labeling process provides ground truth at room-level granularity, it is beneficial to combine $L_{\text{MIDR}}$ and $L_{\text{MSE}}$ into a hybrid loss function via a weighted sum to train the MFSP predictor. In this way, the combined loss, $L_{\text{Hybrid}}$, is defined as follows:  
\begin{equation}
    L_{\text{Hybrid}} = w_1 \, L_{\text{MIDR}} + w_2 \, L_{\text{MSE}},
    \label{eq:hybrid}
\end{equation}  
where $w_1$ and $w_2$ are weights balancing the contributions of the two loss components. In this study, we fix $w_2=1$, so that $w_1$ is interpreted as a measure of the distrust (increasing as $w_1$ increases) in the pseudo ground truth MFSP generated by the labeling process. This hybrid approach combines the strengths of leveraging differentiability through $L_{\text{MIDR}}$ and the guidance provided by pseudo ground truth through $L_{\text{MSE}}$, offering a robust method for training the MFSP predictor.
\begin{figure*}[h!]
    \centering
    \includegraphics[width=0.8\linewidth]{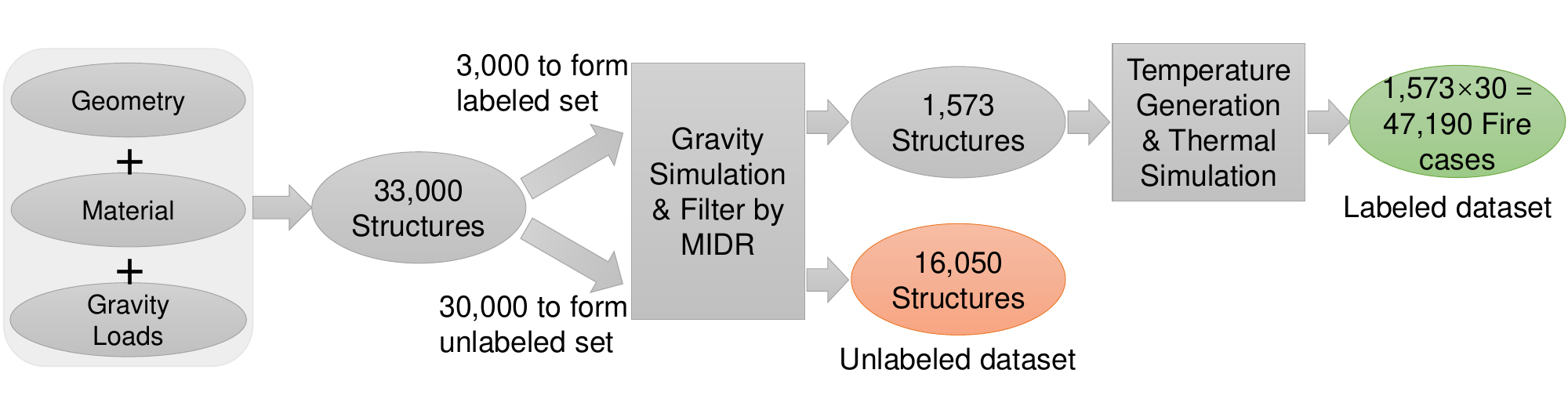}
    \vspace{-3ex}
    \caption{Workflow for dataset generation (geometry, material property, gravity loads, and fire scenarios).}
    \label{fig:dataset_generation_procedure}
\end{figure*}

\subsection{TL with GNN in MIDR Predictor}
\label{subsec:mfspp_transfer_learning}
The MFSP predictor operates as an ``argmaxer'' for the MIDR predictor. While it could be designed as a completely new network and trained from scratch (i.e., De novo training), we propose leveraging the pretrained GNN module from the MIDR predictor through TL. The GNN module in the MIDR predictor captures comprehensive global structural information, making it a suitable platform for the MFSP predictor. Specifically, we reuse the GNN module as a {\em{feature extractor}}, replace the MLP task head with a new one tailored for the MFSP task, and {\em{fine-tune}} the entire model to optimize its performance.
By reusing the pretrained GNN module, which encodes essential structural and gravity load features, the MFSP predictor can focus on learning the relationships necessary to identify the MFSP. This approach reduces training time and enhances model performance by building on the already-learned representations of the structural information, rather than starting from scratch.

A unique challenge arises when reusing the GNN in the MIDR predictor requiring a fire point as the input, which is not need for the MFSP prediction. Common practice to deal with such problem is {\em{masking}}, especially in computer vision or natural language processing. However, masking, such as zero-value masking (i.e., ignore certain inputs by setting them to zero), cannot be directly applied here because the 3D coordinates $(0,0,0)$ represent a valid location --a possible fire at the bottom corner of the building structure. To overcome this challenge, we introduce a randomized Virtual Fire Point (VFP). Essentially, this VFP can be regarded as a placeholder for the GNN input. During training, a random virtual fire location is assigned to each structure for every iteration, independently of prior iterations. This randomized approach forces the network to learn the global structural features necessary for predicting the MFSP, while effectively ignoring the influence of specific fire locations. 
In the inference stage, different from the training stage, we set the VFP to be the geometric center of the structure, i.e., not being randomized. This ensures that the MSFP predictor focuses on the overall structural information rather than being biased by the presence of specific fire locations. By combining TL and the VFP, the MFSP predictor achieves robust generalization and improved accuracy in identifying the MFSP for diverse structural configurations.

\section{Dataset Generation}
\label{sec:dataset}
To train the proposed GNN, we constructed a dataset of building structures and a subset of these structures are subjected to fire simulations using FEA. The preliminary design phase of buildings is the focus of the application of the proposed framework. Therefore, the dataset is viewed as various structural design drafts, where some of these drafts may not only fail to meet the fire requirements, but also fail to satisfy other requirements such as earthquake resistance. The dataset generation process is illustrated in \figref{fig:dataset_generation_procedure}. Initially, a total of 33,000 building structures with geometrical details, material properties, and gravity loads are created. Due to the randomness in generating these structures, a filter is applied to remove unreasonable data after gravity load simulations, which included 15,377 structures. A trade-off between computational feasibility and model performance is made among the remaining 17,623 structures. As further labeling structures with MIDR requires resource-intensive fire simulations via $\chi$ara, a large proportion (16,050 structures) is selected as unlabeled dataset. On the other hand, each of the other 1,573 structures was further subjected to 30 different fire simulations, forming the labeled dataset containing $1,573\times 30 = 47,190$ fire cases. This section details the step-by-step process for generating the dataset, including geometry creation, material property assignment, and simulations due to gravity loads and fire scenarios.

\begin{figure*}[h!]
    \centering
    \begin{subfigure}[b]{0.32\linewidth}
        \centering
        \includegraphics[width=\linewidth]{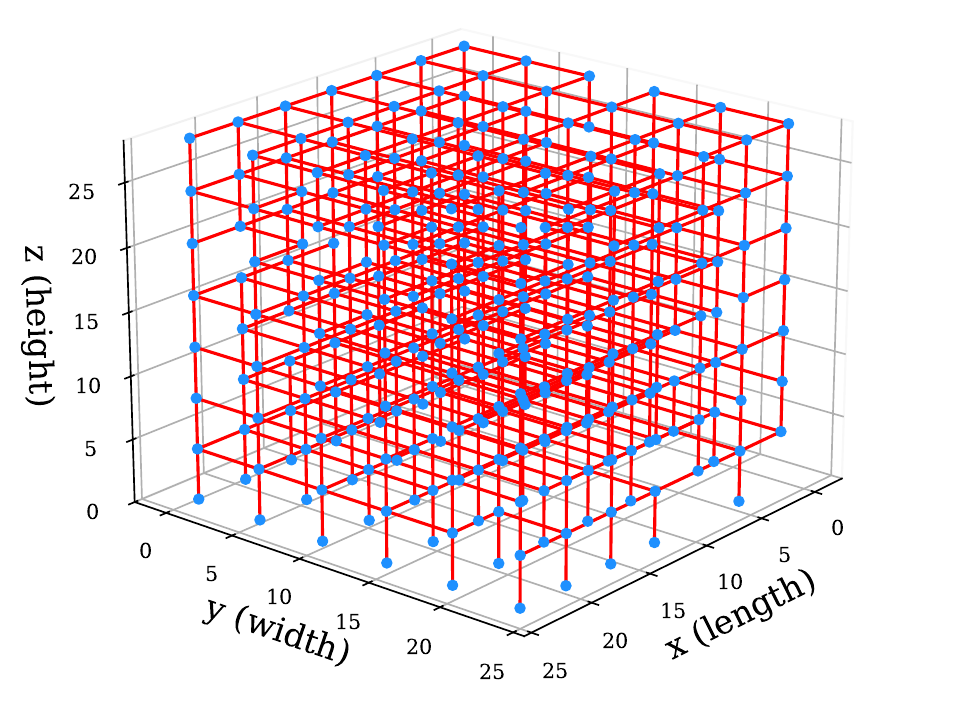}
        \customsubcap{\texttt{ex1}: $22.97 \times 23.30 \times 28.19 $}
        \label{fig:example_generated_geometry_1}
    \end{subfigure}
    \hfill
    \begin{subfigure}[b]{0.32\linewidth}
        \centering
        \includegraphics[width=\linewidth]{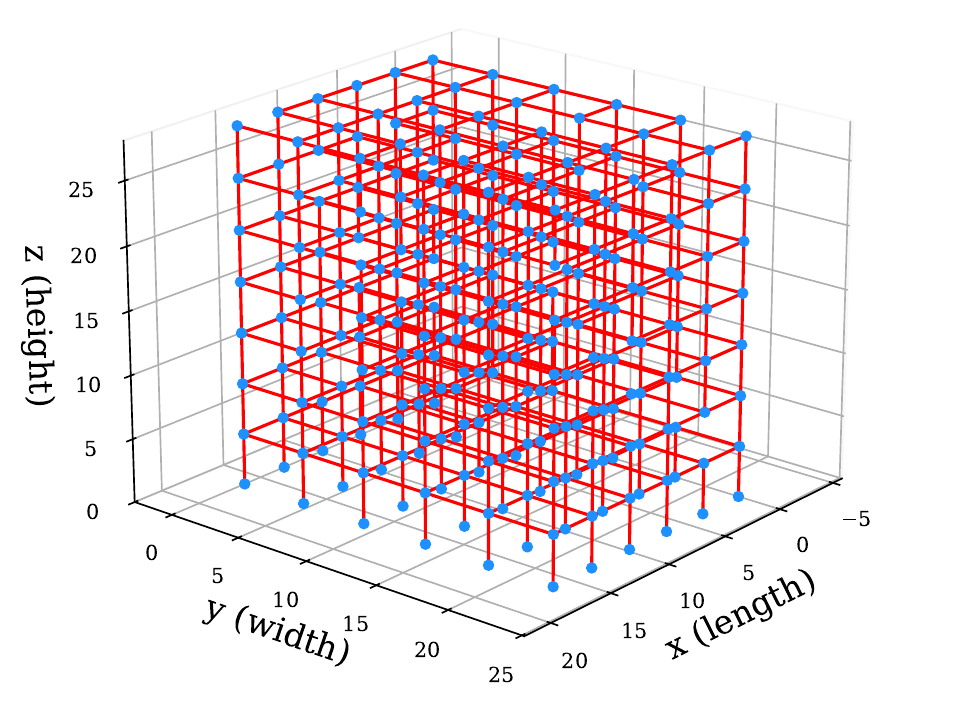}
        \customsubcap{\texttt{ex2}: $16.30 \times 22.30 \times 28.02 $}
        \label{fig:example_generated_geometry_2}
    \end{subfigure}
    \hfill
    \begin{subfigure}[b]{0.32\linewidth}
        \centering
        \includegraphics[width=\linewidth]{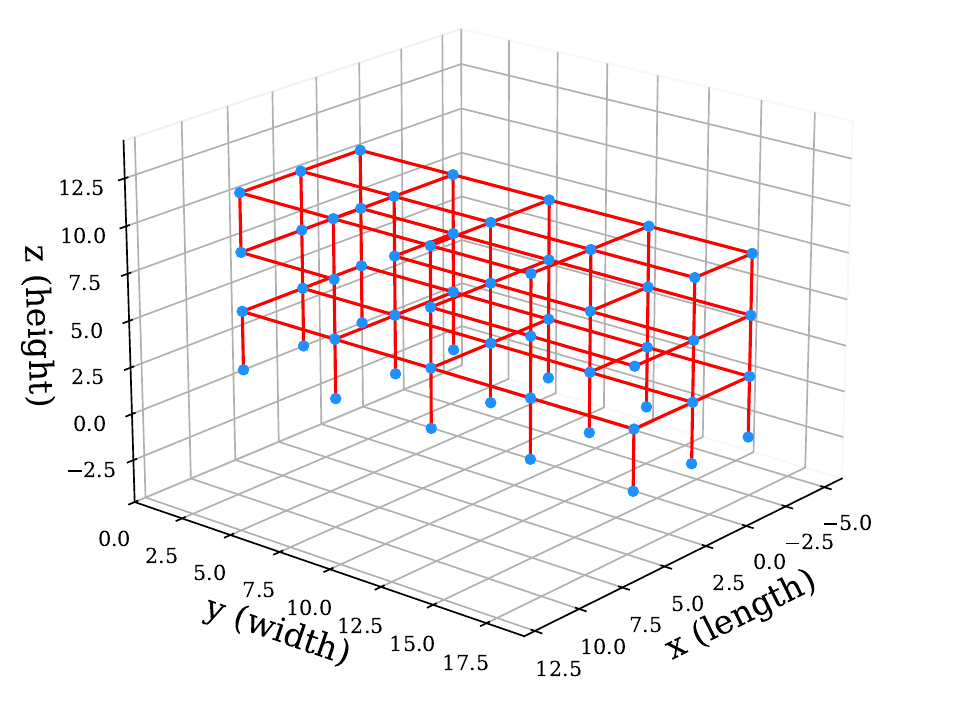}
        \customsubcap{\texttt{ex3}: $6.93 \times  19.25 \times 9.66 $}
        \label{fig:example_generated_geometry_3}
    \end{subfigure}
     \hfill
    \caption{Examples of generated structural geometry of different sizes (all dimensions in meters).}
    \label{fig:example_generated_geometry} 
\vspace{-3ex}
\end{figure*}

\subsection{Geometry Generation}
\label{subsec:geometry_generation}
The geometry of the building structures forms the foundation of the dataset. Regular 
3D buildings resembling multi-story parking structures or shopping malls are generated, with parameters such as building floor dimensions and story heights selected randomly. Each building structure is composed of multiple rooms, where a room serves as the basic unit in this study. A room herein is a cuboid space defined by specific length, width, and height. Within a structure, rooms of the same dimensions are uniformly arranged along the length, width, and height, corresponding to the $x$-, $y$-, and $z$-axes, respectively. Building structures vary in room size and number of rooms along each axis. Specifically, the room length, width, and height are independently sampled from a uniform distribution within the interval $[2, 5]$ meters along the three directions of the structure. Similarly, the room number along each axis is uniformly sampled independently as an integer within the interval $[2, 7]$, i.e., the maximum number of stories of the simulated buildings in this study is 7.

To introduce variability and simulate real-world scenarios, approximately $8\%$ of the structural elements (beams or columns) are randomly removed after initial geometry creation. 
Such removal is not fire-induced damage, but reflects functional diversity often observed in real buildings, such as open spaces designed for activities in shopping malls, e.g., ice skating rinks. Three examples (i.e., \texttt{ex1}, \texttt{ex2} \& \texttt{ex3}) of the generated geometries are illustrated in \figref{fig:example_generated_geometry}, showcasing the diversity and realism of the dataset. This element removal does not affect the definition of room's geometry in the structure and nor does it affect the number of considered fire scenarios.

A range of coefficient of variation values ($3.3\%$ to $17.5\%$) was derived from prior studies that investigated the statistics of geometrical and material properties of structural components of buildings (e.g., \cite{mirza1979variations, lee2004probabilistic}). These studies provide empirical data on the natural variability in parameters such as Young's modulus, yield strength, and dimensions of structural elements due to manufacturing tolerances and material inconsistencies. By selecting $8\%$ for the removal of structural elements in our database, we aimed to maintain a level of variability that is representative of real-world uncertainties while ensuring computational feasibility. This choice ensures that the database captures realistic deviations without introducing extreme cases that may not be commonly encountered in practice.

In this study, we opt for a deterministic square, dimension of $0.1$ m, solid cross-sectional steel elements due to their simplicity in modeling and analysis. Square sections exhibit uniform geometrical properties in all directions, simplifying the computation of structural responses and avoiding complications associated with more complex shapes, such as wide-flange sections, facilitating the computational efficiency and scalability to generate a large dataset. This choice also helps to mitigate issues related to stress concentrations and facilitates a more straightforward representation of structural behavior under thermal loads. 

\textit{Remark:} 
Uniform square cross-sections are adopted for both beams and columns in this study, prioritizing computational efficiency and minimizing the number of design parameters during database generation, rather than strictly following standard capacity design principles. This simplification enables a scalable framework for analyzing the fire-induced response of generic steel frames, with the primary objective being the assessment of fire vulnerability through ML-based predictions. Nonetheless, for scenarios involving additional loads such as seismic or wind actions, it would be necessary to implement larger sections, apply the strong-column/weak-beam design philosophy, and incorporate ductile detailing to realistically represent structural behavior under combined loading effects. Future research could further explore the effects of varying cross-sectional dimensions and semi-rigid connections on structural performance during different fire scenarios.

\subsection{Material Properties}
Steel is chosen as the material for the structures. To reflect real-world variations, different structures are assigned with steel materials with different properties. The ranges of material properties are provided in \tabref{tab:material_property_ranges} and the properties are sampled from uniform distributions of the corresponding ranges. These variations simulate differences arising from manufacturing batches or regional material properties. These ranges are obtained by firstly choosing $210$ GPa, $275$ MPa and $1\%$ as typical values for Young's modulus, yield strength and strain-hardening ratio and then allowing a 20\% variation around these typical values. Note that these properties are at ambient temperature and change when the temperature rises due to fire. The selection of materials with varying properties is aimed at increasing the diversity of the data. Our goal is to represent as wide range of data as possible with a limited amount of building structure data, thereby enhancing the generalization ability of the GNN. Our assumed material property ranges are expected to be wider than the real-world conditions based on findings in \cite{mirza1979variations, lee2004probabilistic}. Therefore, we are essentially tackling a more challenging and general task. If we can solve this problem, we are confident that our method will perform equally well or even better in real-world applications.

\begin{table}[h!]
    \centering
    \caption{Ranges of material properties for considered steel structures.}
    \begin{tabular}{lc}
        \toprule
        Property & Range \\
        \midrule
        Young's modulus & [168, 252] GPa \\
        Yield strength & [220, 330] MPa \\
        Strain-hardening ratio & [0.8, 1.2] \% \\
        \bottomrule
    \end{tabular}
    \label{tab:material_property_ranges}
\end{table}

\subsection{Gravity Loads}
Gravity loads are applied to columns and beams based on their influence (tributary) areas as typically conducted in structural design. The considered load conditions include the column self-weight and the additional loads directly supported on the beams from their self-weight and weights of the reinforced concrete slabs, and building content. An edge beam typically carries approximately half the gravity load supported by a parallel interior beam. The gravity loads are set to be uniformly sampled from ranges to  reflect real-world variability, as listed in \tabref{tab:gravity_load_ranges}. Structures that failed to meet an MIDR threshold of $1\%$ under gravity loads are deemed unacceptable designs and filtered out, as such configurations of randomly chosen geometry, material, and gravity load combinations are considered unrealistic from a regulatory and practicality points of view.
\begin{table}[h!]
    \centering
    \caption{Gravity load ranges for considered beams and columns.}
    \begin{tabular}{lc}
        \toprule
        Element & Range (kN/m)  \\
        \midrule
        Column & [0.5, 1.0]  \\
        Edge beam & [1.5, 4.5]  \\
        Interior beam & [3.0, 7.5]  \\
        \bottomrule
    \end{tabular}
    \label{tab:gravity_load_ranges}
\end{table} 

\subsection{Rule-based Thermal Load Generation}
\label{subsec:thermal_load_generation}
To evaluate a building's structural response during a fire event, we employ a simplified rule-based approach for thermal load generation. 
According to \cite{spearpoint_fire_2008}, a typical fire development follows a predictable pattern. During the {\em{growth stage}} after ignition, the temperature rises slowly and approximately linearly. This is followed by the {\em{burning stage}} starting from the onset of {\em{flashover}}, where temperatures increase rapidly to peak values. After reaching the peak, the temperature either stabilizes or continues to rise slowly until the {\em{decay stage}} begins. Inspired by this fire development pattern, we describe the temperature evolution in time, $t$, prior to the decay stage in two distinct stages:
\begin{enumerate}
    \item {\bf{Initial linear increase stage}}: For $t \in [0, t_1)$, temperature increases gradually and linearly as the fire spreads through the building. This stage represents the time before the fire directly affects a structural element.  
    \item {\bf{ISO 834 fire curve stage}}: For $t \in [t_1, t_{\thre}]$, temperature rises rapidly following the ISO 834 curve \cite{ISO834}, modeling the direct impact of the fire on the structural element. 
\end{enumerate}
The slope of the linear temperature increase, $c$, and the transition time, $t_1$, are influenced by the spatial relationship between the fire source and the structural element. For the second stage of temperature evolution, we utilize the ISO 834 curve, a widely accepted standard for fire resistance testing. This standardized fire curve describes the temperature rise over time, enabling rapid and consistent thermal fields across various scenarios. The duration of fire simulation in this study is set to $t_{\thre}=60$ minutes. This value represents the upper limit for the temperature evolution of each structural element, providing a consistent basis for analyzing the structural response to fire.

Let $(x, y, z)$ represents the midpoint of a structural element and $(x_{\subfire}, y_{\subfire}, z_{\subfire})$ the fire source point. Integer parameters $h$ and $h_{\subfire}$ correspond to the respective floor levels of the element and the fire source. The temperature evolution for each element is expressed as follows:
\begin{enumerate}
    \item Linear increase stage ($0 < t < t_1$):
    \begin{equation}
    T(t) = c \cdot t,
    \end{equation}
    where $c$, the rate of temperature increase ($^\circ\mathrm{C}/\mathrm{min}$), depends on the height difference between the element, $h$, and the fire source, $h_{\subfire}$:
    \begin{equation}
        c = 
        \begin{cases} 
        5\left/\left(h - h_{\subfire} + 1\right)\right., & h \geq h_{\subfire}, \\
        2\left/\left(h_{\subfire} - h\right)\right., & h < h_{\subfire}.
        \end{cases}
    \end{equation}
     \item ISO 834 stage ($t \geq t_1$):
\begin{equation}
    T(t) = c \cdot t_1 + 345 \log_{10} \left(8 \left(t - t_1\right) + 1\right).
\end{equation}
\end{enumerate}

The transition (arrival) time $t_1$ in minutes, i.e., end of the linear stage, depends on the spatial distance between the fire source and the element. We define the following two Euclidean distances $L_p$ in the $X-Y$ plane and $L_s$ in the $XYZ$ space:
\begin{eqnarray}
L_p & \triangleq & \sqrt{(x - x_{\subfire})^2 + (y - y_{\subfire})^2}, \\
\label{eq:Lp}
L_s & \triangleq & \sqrt{(x - x_{\subfire})^2 + (y - y_{\subfire})^2 + (z - z_{\subfire})^2}.
\label{eq:Ls}
\end{eqnarray}
Accordingly, the transition time, $t_1$, is expressed as follows:
\begin{equation}
    t_1 = 
    \begin{cases}
    \beta_{1} \cdot \left(1 - \exp\left\{- L_s\left/\alpha_{1}\right.\right\}\right), & h > h_{\subfire}, \\
    \beta_{2} \cdot \left(1 - \exp\left\{- L_p\left/\alpha_{2}\right.\right\}\right), & h = h_{\subfire}, \\
    \beta_{3} \cdot \left(1 - \exp\left\{- L_s\left/\alpha_{3}\right.\right\}\right), & h < h_{\subfire} .
    \end{cases}
    \label{eq:t1}
\end{equation}
The parameters $\beta_i$ (same time unit as $t_1$) and $\alpha_i$ (same length unit as $L_s$ or $L_p$) for determining $t_1$ are summarized in Table~\ref{tab:fire_spread_parameters}. In this study, we take $r_{\mathrm{up}}=0.95$ and $r_{\mathrm{down}}=0.97$.
\begin{table}[ht]
    \centering
    \caption{Fire spread parameters for $t_1$ calculations.}
    \begin{tabular}{lcc}
        \toprule
        Case  & $\beta_i$ & $\alpha_i$  \\
        \midrule
        $i=1$, Upward spread & $16 \left.\left(1-r_{\mathrm{up}}^{\left|h-h_{\subfire}\right|}\right)\right/\left(1-r_{\mathrm{up}}\right)$ & $10$  \\
        $i=2$, Horizontal spread & $18$ & $18$  \\
        $i=3$, Downward spread & $30 \left.\left(1-r_{\mathrm{down}}^{\left|h-h_{\subfire}\right|}\right)\right/\left(1-r_{\mathrm{down}}\right)$ & $5$  \\
        \bottomrule
    \end{tabular}
    \label{tab:fire_spread_parameters}
\end{table}

\begin{figure}[h!]
    \centering
    \includegraphics[width=0.8\linewidth]{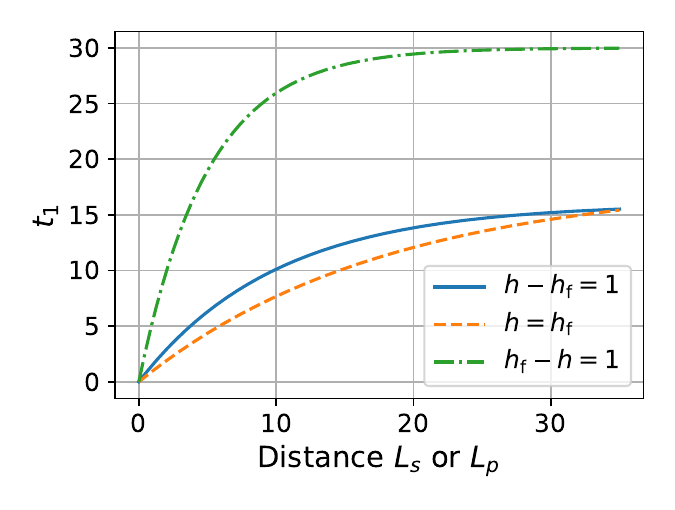}
    \caption{Three examples for the $t_1$ curve.}
    \label{fig:t1_curve}
\end{figure}

\figref{fig:t1_curve} illustrates the $t_1$ curves ($t_1$ in minutes vs. $L_s$ or $L_p$ in meters) for various fire scenarios: (1) fire originating on the lower floor, $h-h_{\subfire}=1$ with rapid upward spread, (2) fire on the same floor, $h=h_{\subfire}$ with the fastest spread, and (3) fire on the upper floor, $h_{\subfire}-h=1$ with slow downward spread. The exponential decay terms in $t_1$ relationships (Equations \ref{eq:t1}) capture the acceleration of fire spread as the distance increases. As shown in \figref{fig:t1_curve}, the adopted simplified model aligns well with the dynamic model based on Markov chains presented in \cite{cheng_dynamic_2011}, wherein rooms on the same floor as the ignition point undergo flashover slightly earlier than those on the floors above. Moreover, both $\beta_1$ and $\beta_3$ can be expressed as the sum of geometric series, with the story level $h$ serving as the index. The fact that the ratios $r_{\mathrm{up}} < 1$ in $\beta_1$ and $r_{\mathrm{down}} < 1$ in $\beta_3$ implies an increasing rate of vertical fire spread across stories, which is consistent with the empirical fire propagation mechanisms discussed in \cite{hokugo_mechanism_2000}. The temperature profile within the range $t \in [0, t_{\thre}]$ is subsequently used as the thermal load in $\chi$ara simulations to compute the displacements at each structural node at time $t_{\thre}$.

\textit{Remark:} The effects of structural elements, such as concrete floor slabs and partitions, are not explicitly modeled in our approach. Instead, their influence is implicitly captured through the careful selection of the parameters $ \alpha_i, \beta_i, r_\mathrm{up} $, and $ r_\mathrm{down} $. This parameterization provides a unified framework for generating temperature fields. Indeed, fire propagation is governed by a multitude of factors and remains an open research question. For instance, if the fire resistance of a floor slab is enhanced by fire protective coating, the corresponding model can account for this by decreasing $\alpha_1$ \& $\alpha_3$, increasing $\beta_1$ \& $\beta_3$, and adopting larger values for $r_\mathrm{up}$ \& $r_\mathrm{down}$, which collectively slows down the vertical spread of fire. Conversely, scenarios involving higher amounts of combustible materials would warrant the opposite adjustments. This flexible and integrated approach avoids the need to design and analyze separate models for different fire propagation scenarios while still capturing the essential effects.

In conclusion, our rule-based approach is a computationally efficient method for approximating fire temperature fields, enabling large-scale dataset generation to train predictive models. By combining ISO 834 fire curves with spatial considerations and embedding structural effects through parameter calibration, the method achieves a balanced trade-off between accuracy and scalability, making it a practical solution for thermal load modeling in fire scenarios. After generating the temperatures of the midpoints of each beam and column, these temperatures are applied as uniform thermal load to these elements of the building structure in question using $\chi$ara. 

\subsection{$\chi$ara Simulation}
\label{subsec:opensees_simulation}

The thermal and mechanical responses of 3D frame structures under combined fire and gravity loads are simulated using $\chi$ara In the simulation, the IDR of each node at $t_{\thre}$ is determined using the computed nodal displacements. Each structural model features 6 Degrees of Freedom (DoFs) per node (3 translations \& 3 rotations), with linear geometrical transformations (\texttt{geomTransf: Linear}) defining how the element local coordinate systems are mapped to the global coordinate system and assuming small displacements and rotations. Although $\chi$ara allows a variety of options for modeling finite deformations, in the present simulations and mainly for simplicity, we did not consider large deformations. All bottom nodes (on the ground) are fully constrained in all 6 DoFs, while those of all other nodes are free. Material behavior is temperature-dependent and modeled with \texttt{Steel01Thermal}, while fiber-based sections (\texttt{FiberThermal}) capture nonlinear interactions between thermal and mechanical responses at the cross-section level. Structural elements are modeled as displacement-based Euler-Bernoulli beam-columns (\texttt{dispBeamColumnThermal}). This element  formulation accounts for thermal strains (temperature gradients) in the section, which is discretized into fibers. Numerical integration is used along the length of each element using three integration (Gauss) points, one at each end and the third in the middle of the element.

Thermal expansion of steel members plays a crucial role in IDR development. In reality, reinforced concrete floor slabs heat at a different rate than steel members due to their higher thermal mass and lower thermal conductivity. This differential heating can lead to restrained thermal expansion, introducing axial compression in beams and affecting the overall structural response. In this study, explicit {\em{composite action}} between steel members and concrete slabs is not modeled. Instead, our approach focuses on isolating the response of the steel structural frame, which is often the critical load-bearing component in fire scenarios. This assumption aligns with prior studies \cite{Possidente_2024} demonstrating that steel structures reach thermal equilibrium with surrounding gases quickly, allowing the use of uniform thermal loading in fire analysis. Future work could enhance this framework by incorporating slab-beam interaction effects, through a refined FEA for an extended dataset where constraints imposed by floor slabs are explicitly considered.

The analysis begins with the application of gravity loads, followed by incremental thermal loads simulating the fire exposure. A static nonlinear solver using  \texttt{ExpressNewton} algorithm ensures convergence, while the \texttt{NormDispIncr} test maintains accuracy. An incremental \texttt{LoadControl} scheme with small step sizes is employed to guarantee numerical stability, using 10\% for gravity loads and 1\% for thermal loads. 

In the thermal load analysis, uniform thermal load is applied to each beam and column, i.e., the temperature of each element is set to be that at the middle point, according to \secref{subsec:thermal_load_generation}. The \texttt{Steel01Thermal} material allows the properties (e.g., Young's modulus and yield strength) to be adjusted at increasing temperatures according to \cite{EN1993} using its Table 3.1: Reduction factors for the stress-strain relationship of carbon steel at elevated temperatures. For example, if Young’s modulus at ambient temperature is $E_0$, then as the temperature $T$ increases, the modulus changes as $E(T) = \eta (T) \times E_0$. Reference \cite{EN1993} directly provides the values of $\eta(T) \in \left[0,1\right]$ at every $100 ^\circ\mathrm{C}$ interval and recommends linear interpolation to obtain $\eta(T)$ for intermediate $T$ values. OpenSees documentation \cite{OpenSeesThermalExamples} provides several examples of thermal analyses.

This modeling framework accommodates variations in material properties, cross-sectional geometries, and temperature profiles, providing robust simulations of structural behavior under fire conditions. The primary settings and configurations for the $\chi$ara simulations are summarized in \tabref{tab:ops_detail}.
\begin{table}[h!]
    \centering
        \caption{Key settings of $\chi$ara simulations.}
    \begin{tabular}{l|>{\raggedright\arraybackslash}p{0.6\linewidth}} %
    \toprule
    Modeling Aspect     & Details \\
    \midrule
    Geometry            & 3D models; 6 DoFs per node \\
    Transformation      & geomTransf: Linear \\ 
    Material            & Steel01Thermal \\
    Section             & FiberThermal; Cross-section: $0.1$ m $\times$ $0.1$ m \\ 
    Element type        & {dispBeamColumnThermal} \\ 
    Loading             & Gravity loads: {beamUniform}; Thermal loads: {beamThermal} \\
    Integration scheme  & Incremental {LoadControl}; Step size: $10\%$ (gravity analysis), $1\%$ (thermal analysis) \\
    Nonlinear solver    & {ExpressNewton} algorithm; {UmfPack} solver; Convergence test: {NormDispIncr} tolerance: $10^{-8}$; Maximum \# iterations per step: $1000$. \\ 
    \bottomrule
    \end{tabular}
    \label{tab:ops_detail}
\end{table}

For each structure in the labeled dataset, 30 fire points are selected using a {\em{dual-granularity}} approach, i.e., two-stage sampling strategy, to ensure they are well-distributed. Specifically, rooms are sequentially selected, with one fire point randomly chosen within each selected room. If a building is large and contains more than 30 rooms, we randomly select 30 rooms without replacement, i.e., ensuring that no more than one fire point is located in the same room. Conversely, if the building is small and has fewer than 30 rooms, all rooms are initially selected, with one fire point randomly assigned to each room. Additionally, rooms are then selected with replacement until a total of 30 fire points are assigned. The room-level sampling prioritizes selecting distinct rooms to avoid spatial clustering of fire points, while the point-level sampling ensures intra-room variability. This approach aligns with stratified sampling principles commonly used for efficient spatial representation, where multi-stage sampling strategies optimize coverage and variability, e.g., \cite{arunachalam_generalized_2023}, and enables a more comprehensive characterization of how the structures respond under fire conditions.

\subsection{Summary of the Dataset Generation}
\begin{figure*}[h!]
    \centering
    \includegraphics[width=\linewidth]{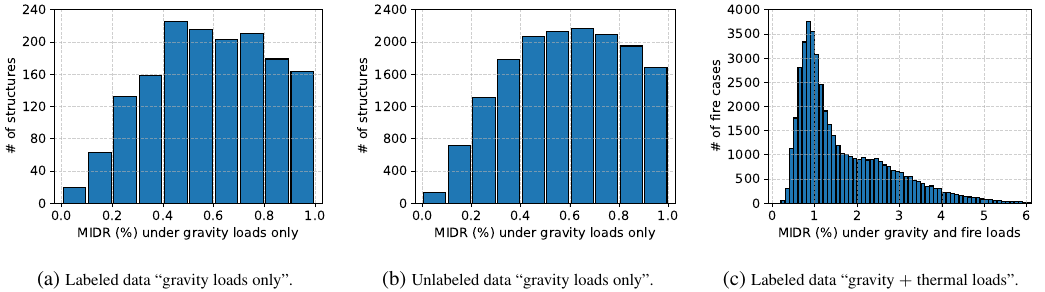}
    \caption{Histograms of MIDR for labeled and unlabeled building structures with gravity and fire load cases.}
    \label{fig:histogram_mdr}
\end{figure*}

As discussed in this section and related to  \figref{fig:dataset_generation_procedure}, three key steps are considered in the development of the dataset: 
\begin{enumerate}
    \item {\bf{Filtering process}}: Structures with MIDR exceeding $1\%$ under gravity loads are excluded,  resulting in $1,573$ labeled structures retained for fire simulations and $16,050$ unlabeled structures for training the MFSP predictor.
    \item {\bf{Fire simulations}}: For each retained labeled structure, 30 fire scenarios are simulated using $\chi$ara ($47,190$ fire cases).
    \item {\bf{Data distribution check}}: MIDR distributions for labeled and unlabeled data under gravity loads are highly similar because both datasets are generated using the same method. Under fire conditions, the MIDR distribution shifted, reflecting significant structural deformation with values reaching a maximum of about 6\%, an average of 1.70\%, and a standard deviation of 1.12\%. This step ensured a diverse and comprehensive dataset for the proposed predictive framework.
\end{enumerate}
The statistical distribution histograms for MIDR (after applying the $1\%$ filtering threshold for gravity load responses) under different loading conditions are plotted in \figref{fig:histogram_mdr}. Figures \ref{fig:histogram_mdr}(a) and \ref{fig:histogram_mdr}(b) show the MIDR distributions of the labeled and unlabeled data, respectively, under gravity loads only. \figref{fig:histogram_mdr}(c) shows the MIDR distribution of the labeled data under the combined effects of gravity and fire loads. Clearly, fire loads cause the structures to significantly deform, leading to a noticeably right-skewed MIDR distribution.

This dataset provides the basis for training and testing the performance of the GNN-based framework. Although we employed a simplified rule-based thermal load generation method compared with conventional CFD-based simulations, the temperature field, the changes of the material properties, and the response of the structures, are all still highly nonlinear and complex. Therefore, it is still a challenging task for the NN to predict the MIDRs based on this dataset.

\textit{Remark}: The dataset generation method involves many simplifications that may not all be representative of the final designs of real-world buildings. However, as stated in \secref{introduction}, the main usage of the MFSP predictor is in the preliminary design stage where several conceptual designs are explored in an iterative manner and these designs may not be initially in complete compliance with all code provisions. Therefore, although the dataset may contain some unrealistic design cases, it can be viewed as the initial conceptual designs to be evaluated and therefore is valid in evaluating the effectiveness of the intended purpose of the proposed framework.
\section{Implementation \& Evaluation}
\label{sec:evaluation}
This section outlines the implementation details and evaluates the performance of both the MIDR and MFSP predictors.

\subsection{MIDR Predictor Evaluation}
\label{subsec:mdrp_eval}

\subsubsection{Implementation details}
To explore the impact of model capacity on performance, we implemente two versions of the MIDR predictor: the {\em{small predictor}} and the {\em{large predictor}}. Both models share an identical network architecture but differ in the dimensions of the embedding vectors used for graph nodes and edges. The architecture and hyper-parameters are chosen with a trial-and-error method and ``min-max'' normalization is utilized for data-scaling. The small and large predictors employ embedding dimensions of $32$ and $64$, respectively. These variations allow for evaluating trade-offs between model complexity, computational efficiency, and predictive accuracy. The detailed configurations of each network component, all utilizing ReLU as the activation function in their MLP layers, are outlined below:
\begin{itemize}
    \item {\bf{Initial encoders for nodes and edges}}: Two distinct single-hidden-layer MLPs serve as encoders for nodes and edges, each tailored to its respective input features. The dimensions of their hidden layers are the same as the output dimension. Consequently, for the small predictor, with both node and edge embedding dimensions set to $32$, the node encoder layers are configured as $[13, 32, 32]$ for the input features, hidden layer, and output layer, respectively. Similarly, the edge encoder layers are $[9, 32, 32]$. Refer to \secref{subsubsec:input_attributes} for the input attribute counts. For the large predictor, the embedding dimensions are increased to $64$, and the layer configurations are adjusted accordingly. This design ensures the encoders effectively process distinct feature sets for nodes and edges while seamlessly integrating with the GNN architecture.
    \item {\bf{Message function}}, $\phi^k(\cdot)$: A single-hidden-layer MLP is used for the message function. Its input is the concatenation of the embeddings of the source node and the edge. The output dimension matches the node embedding dimension. The hidden layer dimension for the small and large predictors is $64$ and $128$, respectively.
    \item {\bf{Aggregation operation}}, $\bigoplus$: The ``max'' function is used as the aggregation operation. It selects the maximum value from the neighboring node messages. This approach (compared to other aggregation methods through averaging or summation) aligns with a conservative engineering design philosophy, prioritizing the most critical structural information for robust analysis and decision-making.
    \item {\bf{Update function}}, $\gamma^k(\cdot)$: It combines the previous node embedding with the processed aggregated message. Specifically, the aggregated message for node $i$, $\vec{\tilde{v}}_{i}^k$, is processed through a single-hidden-layer MLP, $\tilde{\gamma}^k(\cdot)$, whose output dimension matches the node embedding dimension. The update rule is as follows: 
    \begin{equation}
        \gamma^k\left(\vec{v}_{i}^{k-1}, \vec{\tilde{v}}_{i}^k\right) = \vec{v}_{i}^{k-1} + \tilde{\gamma}^k\left(\vec{\tilde{v}}_{i}^k\right).
    \end{equation}
    The hidden layer dimension of $\tilde{\gamma}^k(\cdot)$ for the small and large predictors is $64$ and $128$, respectively.
    \item {\bf{EU function}}, $\zeta\left(\cdot\right)$: A single-hidden-layer MLP is used for updating edge attributes. The input is the concatenation of the two adjacent nodes. Thus, the input dimension is twice the node embedding dimension. The output dimension corresponds to the edge embedding dimension. The hidden layer dimension for the small and large predictors is set as $32$ and $64$, respectively.
    \item {\bf{Task head-1, task head-2 \& pooling function}}: Task heads are implemented as {\em{linear}} layers. The pooling function combines node embedding into a graph embedding by concatenating the {\em{mean}} and {\em{max}} pooling results. Consequently, the input dimension for task head-1 (for Strawman 2) is twice the node embedding dimension, while for task head-2 (for Strawman 1), it equals the node embedding dimension. Recall that task head-1 and task head-2 predict the IDR as a node-level outcome and MIDR as a structure-level outcome, respectively, i.e., a scalar value is output as the final prediction for each head.
\end{itemize}

The number of layers in the GNN is aligned with the number of stories in the building structure. As the maximum number of stories of the considered buildings in this study is 7 (refer to \secref{subsec:geometry_generation}), we set the GNN to possess 7 layers. Note that to process a $k$-story structure, only first $k$ layers of the entire GNN is used (refer to \secref{subsubsec:handle_structural_data}). For the small predictor configuration, the total parameter count is approximately $7.1 \times 10^4$, while for the large predictor configuration, it reaches about $2.8 \times 10^5$. In the absence of the EU functionality, the parameter counts are marginally reduced to approximately $6.8\times 10^4$ for the small predictor and $2.7 \times 10^5$ for the large predictor.

\subsubsection{Evaluation metrics}
We evaluate the performance of the MIDR predictor using three metrics: Mean Squared Error (MSE), Mean Absolute Error (MAE), and Spearman's rank correlation coefficient. MSE and MAE are respectively defined as follows:
\begin{equation}
    \text{MSE} = \frac{1}{\sum_{m=1}^{M}N_{m}} \sum_{m=1}^{M} \sum_{n=1}^{N_{m}} \left( \widehat{d}_{m,n, \max} - d_{m,n, \max}^{\, \gt} \right)^2,
\end{equation}
\begin{equation}
    \text{MAE} = \frac{1}{\sum_{m=1}^{M}N_{m}} \sum_{m=1}^{M} \sum_{n=1}^{N_{m}} \left| \widehat{d}_{m,n, \max} - d_{m,n, \max}^{\, \gt} \right|,
\end{equation}

\noindent where $d_{m,n, \max}^{\, \gt}$ is the ground truth MIDR of structure $m$ under fire scenario $n$, obtained from $\chi$ara, and $\widehat{d}_{m,n, \max}$ is the corresponding prediction from the MIDR predictor.

Since our ultimate goal is to train an argmaxer to identify the MFSP, rather than merely predicting the MIDR values, accurately assessing the relative ranking of MIDRs across different fire scenarios for each structure is critical. MSE and MAE can be good metrics for characterizing the closeness between the predicted MIDRs and the ground truth MIDRs. However, they do not adequately capture the ranking performance. Therefore, we introduce the Spearman's rank correlation coefficient, $\rho_s$, which measures the {\em{monotonic relationship}} between two random variables. Formally, let $A$ and $B$ be two random variables, and let $R_{A}$ and $R_{B}$ denote their corresponding rank-transformed values based on the observed samples. The Spearman’s rank correlation coefficient is defined as:
\begin{equation}
    \rho_s\left(A,B\right) \triangleq \rho\left( R_{A}, R_{B} \right) = \frac{\text{cov}\left( R_{A}, R_{B} \right)}{\sigma_{R_{A}}\sigma_{R_{B}}},
    \label{eq:Spearman}
\end{equation}
where $\rho(\cdot,\cdot)$ is the Pearson correlation coefficient, $\text{cov}\left(R_{A}, R_{B}\right)$ is the covariance between the ranked values, and $\sigma_{R_{A}}$ \& $\sigma_{R_{B}}$ denote their respective standard deviations. Unlike Pearson's correlation, which measures linear relationships, Spearman’s correlation assesses how well the orderings of the data match, making it especially suitable for ordinal data or for relationships that are monotonic but not necessarily linear. The range of $\rho_s$ is $[-1, +1]$, with values close to $+1$, $0$, and $-1$ respectively indicating strong positive, weak or no, and strong negative monotonic associations.

In this paper, for structure $m$ with $N_m$ fire scenarios, we treat the ground truth and predicted MIDRs as realizations of underlying random variables. Specifically, we define the ground truth sample as $A\equiv\left\{d_{m, 1, \max}^{\, \gt}, d_{m, 2, \max}^{\, \gt}, \ldots, d_{m, N_m, \max}^{\, \gt}\right\}$, obtained from $\chi$ara simulations and the prediction samples as $B\equiv\left\{\widehat{d}_{m, 1, \max}, \widehat{d}_{m, 2, \max}, \ldots, \widehat{d}_{m, N_m, \max}\right\}$, produced by the MIDR predictor.
The final evaluation metric is the average $\rho_s$ across all structures, providing a robust measure of the prediction quality for the MFSP identification.

\subsubsection{Evaluation results}
For training and evaluating the MIDR predictor, we exclusively use the labeled dataset, splitting it into $80\%$ for training and $20\%$ for testing. Various GNN configurations and training methods are compared, with the results summarized in \tabref{tab:mdrp_eval}.
\begin{table}[h!]
    \centering
    \caption{Evaluation results of the MIDR predictor (numbers in boldface indicate best results).}
    \begin{tabular}{lcccc}
        \toprule
        Method & Network & MAE & MSE & $\rho_s$  \\
        \midrule
        Strawman 1 & \multirow{5}{*}{Small} & $1.017$ & $1.707$ & $0.162$ \\  
        Strawman 2: No EU &  & $0.320$ & $0.232$ & $0.670$ \\  
        Strawman 2: EU &  & $0.314$ & $0.203$ & $0.689$ \\  
        Proposed: No EU &  & $0.300$ & $0.201$ & $0.685$ \\  
        \textbf{Proposed: EU} &  & $\mathbf{0.278}$ & $\mathbf{0.170}$ & $\mathbf{0.725}$ \\  
        \midrule
        Strawman 1 & \multirow{5}{*}{Large} & $0.305$ & $0.202$ & $0.649$ \\ 
        Strawman 2: No EU &  & $0.305$ & $0.201$ & $0.678$ \\ 
        Strawman 2: EU &  & $0.317$ & $0.216$ & $0.704$ \\ 
        Proposed: No EU &  & $0.289$ & $0.175$ & $0.707$ \\ 
        \textbf{Proposed: EU} &  & $\mathbf{0.272}$ & $\mathbf{0.169}$ & $\mathbf{0.742}$ \\ 
        \bottomrule
    \end{tabular}
    \label{tab:mdrp_eval}
\end{table}

\paragraph{Overall accuracy performance analysis}
In \tabref{tab:mdrp_eval}, the methods Strawman 1, Strawman 2, and Proposed are described in \secref{subsec:transfer_learning} and illustrated in \figref{fig:transfer_learning}. For Strawman 2 and Proposed methods, we further assess the impact of enabling the EU functionality. Two key observations are summarized below:
\begin{itemize}
    \item {\bf{EU functionality}}: Comparisons between cases with and without EU reveal that enabling EU consistently enhances model performance. Specifically, the inclusion of EU increases $\rho_s$ in all cases, and reduces both MAE and MSE for not only the proposed predictor but also the Strawman 2 predictor based on small network.
    \item {\bf{Proposed method}}: Among all configurations, the proposed method with EU achieves the best performance, with $\rho_s$ scores of $0.725$ and $0.742$ for the small and large networks, respectively. This indicates that the proposed method better captures the structural sensitivity ranking across various fire scenarios when predicting the MIDR.
\end{itemize}

In structure design, for fire-sensitive buildings, it is expected that the location of MFSP should be predicted more accurately. On the contrary, if a building is not likely to collapse no matter where a fire occurs, then the tolerance for the MFSP prediction errors is higher. For this reason, we specifically consider those buildings whose ground truth MIDR exceeds $2\%$ in at least one fire case, and denote their  data as severe cases. 

\begin{figure*}[h!] 
    \centering
    \begin{subfigure}[b]{0.32\linewidth}
        \centering
        \includegraphics[width=\linewidth]{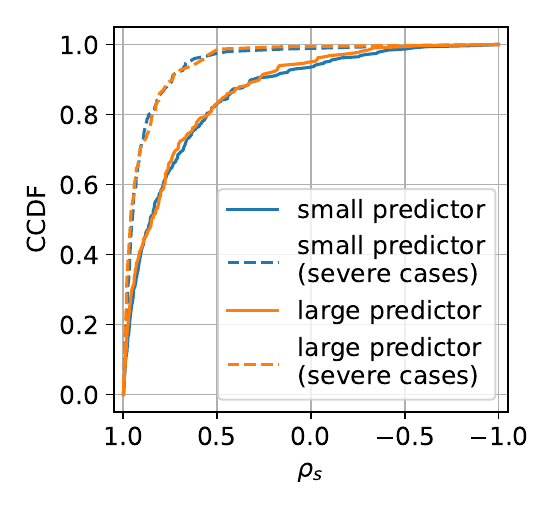}
        \customsubcap{CCDF results ($\rho_s$-axis inverted).}
        \label{fig:midrp_cdf_rho_s}
    \end{subfigure}
    \hfill 
    \begin{subfigure}[b]{0.32\linewidth}
        \centering
        \includegraphics[width=\linewidth]{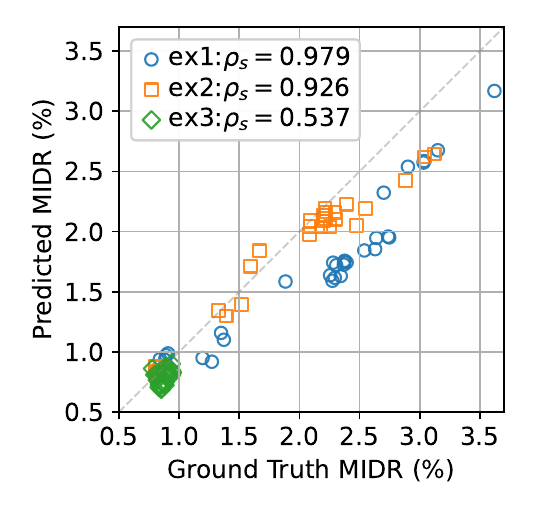}
        \customsubcap{Small predictor.}
        \label{fig:midrp_example_small}
    \end{subfigure}
    \hfill
    \begin{subfigure}[b]{0.32\linewidth}
        \centering
        \includegraphics[width=\linewidth]{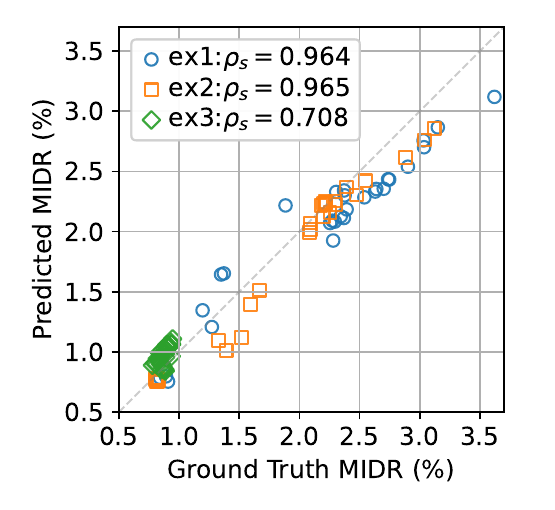}
        \customsubcap{Large predictor.}
        \label{fig:midrp_example_large}
    \end{subfigure}
    \caption{Representative MIDR predictor results: (a) CDF of $\rho_s$ for the proposed MIDR predictors for two different sizes (severe cases are for structures whose ground truth MIDR $>2\%$ in at least one fire case); (b) and (c) show the MIDR predictions vs. the ground truth for 3 example structures (\texttt{ex1}, \texttt{ex2} \& \texttt{ex3}) for the small and large predictors, respectively.}
    \label{fig:midrp_cdf_rho_s_and_examples}
\end{figure*}

We divide the data into severe cases and non-severe cases. In the former, the average variance of the MIDR of each building in the 30 fire scenarios reaches $0.40$, while this value is only $0.02$ in the latter. In other words, for severe cases, MIDRs vary greatly with the location of the fire points. In such cases, $\rho_s$ is an important indicator to measure the reliability of using the MIDR predictor to train the MFSP predictor, and results show that the average $\rho_s$ for severe cases reaches over $0.90$. \figref{fig:midrp_cdf_rho_s_and_examples}(a) presents the Complementary Cumulative Distribution Function (CCDF) of $\rho_s$, i.e., ${\overline{{\cal{F}}}}_{\rho_s}(a)={\cal{P}}(\rho_s \ge a)$, where $\cal{P}$ indicates the probability, for the proposed method with EU. The solid lines include both severe and non-severe cases, i.e., all the test data, where $83.2\%$ of structures achieve $\rho_s > 0.50$, and using small and large predictors $64.4\%$ and $66.7\%$ of structures, respectively, achieve $\rho_s > 0.75$. Additionally, the large predictor improves on the worst case, where the smallest $\rho_s$ increases from $-0.86$ for the small predictor to $-0.58$ for the large predictor. On the other hand, the dashed lines are for the severe cases where for the small predictor, $98.0\%$ of structures achieve $\rho_s > 0.50$ and $89.3\%$ exceed $0.75$ with an average of $0.90$. For the large predictor, $98.7\%$ of structures achieve $\rho_s > 0.50$, while $90.0\%$ exceed $0.75$ with an average of $0.91$.

For the non-severe cases, low MSE of predicted MIDRs guarantees the reliability of the MIDR predictors. Results show that the small predictor leads to an MSE as low as $0.06$ and the value for the large predictor is $0.05$. In the following, the differences between severe and non-severe cases are discussed in detail with three example structures.

\paragraph{Exemplary illustration \& time efficiency}
Here, we present the predicted MIDR vs. the ground truth MIDR for three example structures (i.e., \texttt{ex1}, \texttt{ex2} \& \texttt{ex3} in \figref{fig:example_generated_geometry}) under the small (\figref{fig:midrp_cdf_rho_s_and_examples}(b)) and large (\figref{fig:midrp_cdf_rho_s_and_examples}(c)) predictors. In these examples, \texttt{ex1} and \texttt{ex2} represent severe cases, while \texttt{ex3} remains relatively stable with low MIDR across all fire scenarios, i.e., a non-severe case. As shown in the figures, the predicted results for \texttt{ex1} and \texttt{ex2} closely match the ground truth for both predictors, achieving $\rho_s > 0.92$. For \texttt{ex3}, the $\rho_s$ of both predictors is lower, but the large predictor outperforms the small one, achieving $\rho_s > 0.70$. Although $\rho_s$ for \texttt{ex3} is relatively low, the performance remains accurate, as the predicted and ground truth MIDR values are closely aligned along the ideal $45^\circ$ line with and the MSE $\leq 0.01$. Note that Spearman's rank correlation coefficient does not evaluate the closeness between predicted and ground truth values, but rather evaluates the monotonic relationship. Therefore, in the consequent task of training an argmaxer, the high value of average $\rho_s\geq 0.90$ guarantees the reliability and validity. Moreover, for the non-severe cases, the low MSE indicates that the MIDR predictor is a good surrogate model of thermal simulators.

The differentiable agent shows great improvement on the time efficiency compared with traditional FEA. On our experimental machine, with 2 cores of Skylake CPU (2.1 GHz) and one GTX2080TI GPU, the average time consumptions for predicting the MIDR for the 3 example structures is listed in \tabref{tab:midrp_time}, where the number of nodes and edges are also listed for reference. Note that the values for $\chi$ara are the time consumptions for just one fire scenario, and the values for the agent are those for all 30 fire scenarios. We clearly see that the conventional FEA simulation's efficiency is highly affected by the number of nodes and edges, while these numbers have less impact on the agent's time consumptions. The most important finding, as expected, is that the GNN-based agent is much faster than the FEA simulations, with more than 3 orders of magnitude difference, i.e., $\chi$ara costs several to tens of seconds for only one fire scenario, while the agents cost only several milliseconds (ms) for a batch of 30 fire scenarios. Note that to obtain accurate time consumption of the agent, we run the agent for 1,000 times and report their average run time.
\begin{table}[h!]
    \centering
    \caption{Average time consumptions of different methods to obtain the MIDR for 3 examples in Figures \ref{fig:midrp_cdf_rho_s_and_examples}(b) \& (c).}
    \begin{tabular}{c|cc|c|cc}
        \toprule
        \multirow{2}{*}{\texttt{ex}} & \multicolumn{2}{c|}{Number of} & $\chi$ara & \multicolumn{2}{c}{Agent (1 batch of 30 fires)} \\ 
        \cmidrule(lr){5-6}
        & Nodes & Edges & (1 fire) (ms) & Small (ms) & Large (ms) \\
        \midrule
        \texttt{1} & 331 & 728 & 70,950 & 8.63 & 12.64  \\  
        \texttt{2} & 286 & 618 & 48,340 & 8.59 & 11.00  \\  
        \texttt{3} & 59  & 101 & 6,590  & 4.12 & 4.29 \\  
        \bottomrule
    \end{tabular}
    \label{tab:midrp_time}
\end{table}

In conclusion, the proposed method with EU functionality demonstrates exceptional performance across all metrics for both accuracy and time efficiency, particularly for severe cases. These findings highlight the robustness and effectiveness of the method in supporting fire safety assessments for building structures. The high values of $\rho_s$, and the extremely low time costs also demonstrate the reliability and the efficiency of using the MIDR predictor as an agent for pseudo-labeling and as a direct indicator for training the MFSP predictor. 

\subsection{MFSP Predictor Evaluation}
\label{subsec:mfsp_eval}
To evaluate the MFSP predictor, we adopt the proposed method with EU as the final model for the MIDR predictor and utilize it to train the MFSP predictor. As described in \secref{subsec:mfspp_transfer_learning}, we consider two training strategies:
\begin{itemize}
    \item {\bf{De novo training}} of a new network from scratch.
    \item {\bf{TL with GNN}} by reusing the GNN module, including the initial encoders, from the MIDR predictor. 
\end{itemize}
Additionally, two loss functions are evaluated: 
\begin{itemize}
    \item {\bf{MSE loss}} focuses on the accuracy of predicted coordinates.
    \item {\bf{Hybrid loss}} combines the MSE loss with the MIDR-based loss for better ranking consistency.
\end{itemize}
The network architecture of the MFSP predictor is almost identical to that of the MIDR predictor, consisting of an initial encoder, a GNN, pooling layers, and an MLP task head. There are only a few differences as follows:
\begin{itemize}
    \item {\bf{Final MLP head}}: The task head in the MFSP predictor replaces the linear layer in the MIDR predictor with a single-hidden-layer MLP. The hidden layer's dimension matches the graph embedding (twice the node embedding dimension), and the output layer has a dimension of 3, representing the predicted coordinates $(x, y, z)$. A ``sigmoid'' activation function is applied to normalize the predicted coordinates to the range $[0,1]$, aligned with the building normalized spatial dimensions.
    \item {\bf{De Novo training}}: The initial node input features include only the spatial coordinates of the nodes. Fire point information is excluded, as no prior knowledge from the MIDR predictor is utilized in this case.
    \item {\bf{TL with GNN training}}: The GNN module and initial encoders from the MIDR predictor are directly reused. Moreover, the fire point information is replaced with a VFP to ensure consistency in the input format. This enables the network to generalize its understanding of the structural information without relying on specific fire locations.
\end{itemize}
By evaluating these configurations and loss functions, the MFSP predictor aims to efficiently and accurately predict the MFSP within the building at hand.

\subsubsection{Evaluation metrics}
\label{sec:7.2.1}
In \secref{subsec:mfspp_pseudo_label}, we introduce the ``pseudo-labeling'' method, which generates pseudo ground truth coordinates for the MFSP using the well-trained MIDR predictor. These pseudo-labels are used to evaluate the MFSP predictor through various metrics, capturing both point-level and room-level performances. Specifically, they are generated by selecting the fire point with the highest predicted MIDR for each structure, and subsequently used to train and evaluate the MFSP predictor. This approach allows leveraging the unlabeled data effectively, as the MIDR predictor provides a computationally efficient and accurate surrogate for the FEA simulations.

For a given structure $m$, the distance error between predicted MFSP and the pseudo ground truth MFSP is defined as: 
\begin{equation}
    e_{m} = \sqrt{\left( x_{m}^{\, \pgt} - \widehat{x}_{m} \right)^2 + \left( y_{m}^{\, \pgt} - \widehat{y}_{m} \right)^2 + \left( z_{m}^{\, \pgt} - \widehat{z}_{m} \right)^2},
\end{equation}
where $\widehat{x}_{m}$, $\widehat{y}_{m}$, $\widehat{z}_{m}$ represent the $x$, $y$, $z$ coordinates predicted by the MFSP predictor, while $x_{m}^{\, \pgt}$, $y_{m}^{\, \pgt}$, $z_{m}^{\, \pgt}$ denote the pseudo ground truth MFSP coordinates obtained using the MIDR predictor. This metric provides a direct measure of the prediction accuracy at the point level. However, as the pseudo ground truth is defined at a room-level granularity, its reliability may vary depending on the building's size and complexity. For practical evaluations, two room-level metrics are introduced, namely, room distance error and room rank. These metrics treat rooms as the basic unit of measurements, reflecting how well the MFSP predictor identifies the Most Fire-Sensitive Room (MFSR), i.e., the room where the MFSP is located.

{\bf{Room distance error}} measures the distance between the predicted MFSR and the pseudo ground truth MFSR, assuming all edges (beams and columns) of the building structure have unit length. For structure $m$, let the room dimensions be $\tilde{x}_{m}$, $\tilde{y}_{m}$, $\tilde{z}_{m}$ for the respective length, width, and height. The coordinates of the center point of the predicted MFSR are denoted as $\left( \widehat{x}_{m}', \widehat{y}_{m}', \widehat{z}_{m}'\right)$. The room distance error is then computed as:
\begin{equation}
    \tilde{e}_{m} = \sqrt{\left( \frac{x_{m}^{\, \pgt} - \widehat{x}_{m}'}{\tilde{x}_{m}} \right)^2 + \left( \frac{y_{m}^{\, \pgt} - \widehat{y}_{m}'}{\tilde{y}_{m}} \right)^2 + \left( \frac{z_{m}^{\, \pgt} - \widehat{z}_{m}'}{\tilde{z}_{m}} \right)^2}.
\end{equation}
This metric provides a scaled distance error measure, accounting for the room size and the structural dimensions.

{\bf{Room rank}} evaluates the ranking of the room containing the predicted MFSP based on the pseudo-labeled MIDR values. For structure $m$ with $N_{m}$ pseudo fire cases, we sort the pseudo-labeled room-center MIDRs in a descending order, and index the rooms by the fire case occurring at each room's center, i.e., re-number the rooms to satisfy $d_{m,1, \max}^{\, \pgt} > d_{m,2, \max}^{\, \pgt} > \cdots > d_{m,N_{m}, \max}^{\, \pgt}$. If the predicted MFSP falls into the $l$-th ranked room, the room rank is accordingly defined as $r_{m} = l$. Lower ranks indicate better prediction performance, as the predicted MFSR with such low rank is closer to the MFSP.

In addition to the distance error, room distance error, and room rank, the predicted MIDR at the identified MFSP can also serve as an evaluation metric. Higher MIDR values reflect the predictor's ability to identify the MFSPs, directly correlating with the severity of the structural response. These evaluation metrics collectively assess the MFSP predictor's performance, balancing point-level precision, room-level granularity, and the functional implications of the predictions. The inclusion of room distance error and room rank ensures that the evaluation aligns with practical building-level assessments, where rooms are typically used as reference units for structural analysis and fire safety planning.
\begin{table*}[h!]
    \centering
    \caption{Evaluation results of the MFSP predictor for different training methods and loss functions. Numbers in boldface indicate best results; values in parentheses show performance multiples (indicated with ``$\times$'') relative to the reference configuration, i.e., TL with GNN, Hybrid ($w_1=50$).}
    \begin{tabular}{lcccccc}
        \toprule
        Method & Loss & Network  & Avg. distance error (m) & Avg. room distance error & Avg. room rank & Avg. MIDR (\%)  \\
        \midrule
        De novo training  & MSE               & \multirow{8}{*}{Small} & $5.49~(0.93\times)$ & $1.28~(1.08\times)$ & $9.09~(1.14\times)$ & $2.40~(0.98\times)$ \\ 
        De novo training  & Hybrid ($w_1=10$) &                        & $5.65~(0.96\times)$ & $1.25~(1.06\times)$ & $9.11~(1.15\times)$ & $2.42~(0.98\times)$ \\ 
        De novo training  & Hybrid ($w_1=50$) &                        & $6.10~(1.04\times)$ & $1.24~(1.05\times)$ & $8.50~(1.07\times)$ & $2.45~(1.00\times)$ \\ 
        De novo training  & Hybrid ($w_1=100$)&                        & $6.26~(1.06\times)$ & $1.23~(1.04\times)$ & $8.59~(1.08\times)$ & $2.42~(0.99\times)$ \\ 
        TL with GNN       & MSE               &                        & $\mathbf{5.18~(0.88\times)}$ & $1.21~(1.03\times)$ & $8.51~(1.07\times)$ & $2.40~(0.98\times)$ \\ 
        TL with GNN       & Hybrid ($w_1=10$) &                        & $5.31~(0.90\times)$ & $\mathbf{1.18~(1.00\times)}$ & $8.19~(1.03\times)$ & $2.43~(0.99\times)$ \\ 
        TL with GNN       & Hybrid ($w_1=50$) &                        & $5.89~(1.00\times)$ & $\mathbf{1.18~(1.00\times)}$ & $\mathbf{7.95~(1.00\times)}$ & $\mathbf{2.46~(1.00\times)}$ \\ 
        TL with GNN       & Hybrid ($w_1=100$)&                        & $6.16~(1.05\times)$ & $1.21~(1.03\times)$ & $\mathbf{7.95~(1.00\times)}$ & $2.45~(1.00\times)$ \\ 
        \midrule
        De novo training  & MSE               & \multirow{8}{*}{Large} & $5.41~(0.92\times)$ & $1.27~(1.10\times)$ & $8.02~(1.19\times)$ & $2.37~(0.97\times)$ \\ 
        De novo training  & Hybrid ($w_1=10$) &                        & $5.64~(0.96\times)$ & $1.25~(1.09\times)$ & $7.58~(1.13\times)$ & $2.40~(0.98\times)$ \\ 
        De novo training  & Hybrid ($w_1=50$) &                        & $6.21~(1.06\times)$ & $1.22~(1.06\times)$ & $6.96~(1.03\times)$ & $2.43~(0.99\times)$ \\ 
        De novo training  & Hybrid ($w_1=100$)&                        & $6.46~(1.10\times)$ & $1.23~(1.07\times)$ & $7.27~(1.08\times)$ & $2.42~(0.99\times)$ \\ 
        TL with GNN       & MSE               &                        & $\mathbf{4.93~(0.84\times)}$ & $\mathbf{1.15~(1.00\times)}$ & $7.18~(1.07\times)$ & $2.40~(0.98\times)$ \\ 
        TL with GNN       & Hybrid ($w_1=10$) &                        & $5.21~(0.89\times)$ & $\mathbf{1.15~(1.00\times)}$ & $7.12~(1.06\times)$ & $2.41~(0.98\times)$ \\ 
        TL with GNN       & Hybrid ($w_1=50$) &                        & $5.87~(1.00\times)$ & $\mathbf{1.15~(1.00\times)}$ & $\mathbf{6.73~(1.00\times)}$ & $\mathbf{2.45~(1.00\times)}$ \\ 
        TL with GNN       & Hybrid ($w_1=100$)&                        & $6.23~(1.06\times)$ & $1.18~(1.03\times)$ & $7.20~(1.07\times)$ & $2.43~(0.99\times)$ \\ 
        \bottomrule
    \end{tabular}
    \label{tab:mfspp_eval}
\end{table*}
\subsubsection{Evaluation results}
Using the metrics defined in the previous section, we evaluate the performance of the MFSP predictor by calculating average quantities across all building structures. Since the MIDR predictor provides a computationally efficient agent model to rapidly generate pseudo labels for unlabeled data, the large unlabeled dataset can be effectively utilized for training the MFSP predictor. For this evaluation, the unlabeled dataset is divided into $80\%$ for training and $20\%$ for testing. The evaluation results for various training methods and loss functions are summarized in \tabref{tab:mfspp_eval}. For the ``Hybrid loss'', Equation (\ref{eq:hybrid}), the weight for the MSE term is fixed at $w_2 = 1 \, \mathrm{m}^{-2}$, while the weight for the MIDR term is varied as $w_1=10$, $50$, or $100$. In \tabref{tab:mfspp_eval}, the TL with GNN, Hybrid results for $w_1=50$, using both small and large networks, are taken as the reference cases for the listed obtained multipliers $\mu$ in parentheses, i.e., $(\mu \times)$.

\begin{figure*}[h!]
    \includegraphics[width=\linewidth]{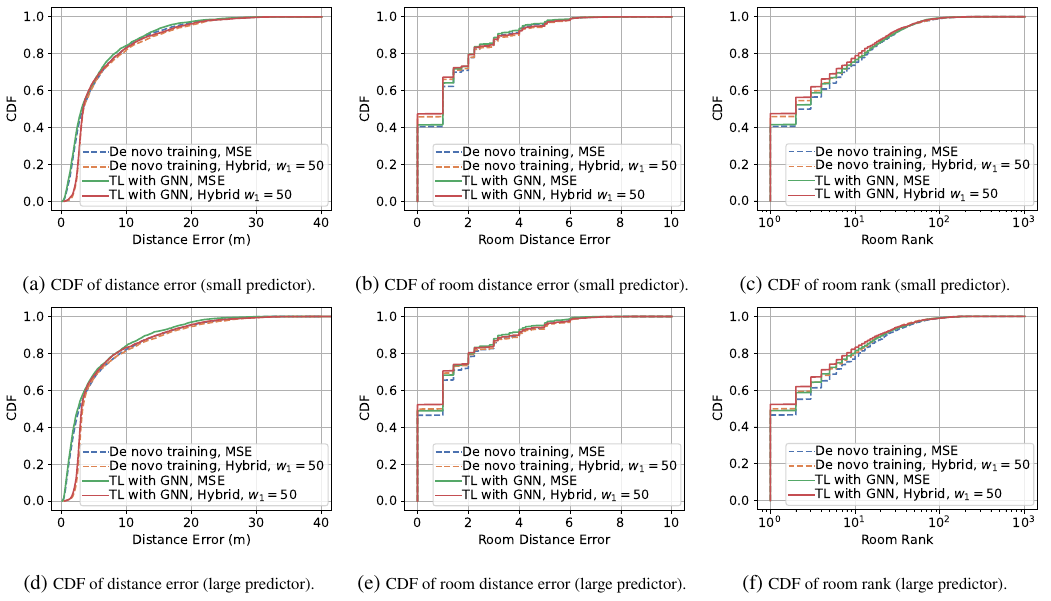}
    \caption{CDF of distance error, room distance error, and room rank for the small (a to c) and large (d to f) MFSP predictors.}
    \label{fig:eval_mfspp_cdf} 
\end{figure*}

\begin{table*}[ht]
    \centering
    \caption{Key CDF values (\%) of different metrics to evaluate the MFSP predictor (numbers in boldface indicate best results).}
    \begin{tabular}{lcccccccc}
        \toprule
        Method & Loss & Network  & $e_{m} \leq 5$ m & $e_{m} \leq 10$ m & $\tilde{e}_{m}\leq \sqrt{2}$ & $\tilde{e}_{m}\leq 2$ &  $r_{m} \leq 5$ & $r_{m}\leq 10$ \\
        \midrule
        De novo training  & MSE               & \multirow{4}{*}{Small} & $63.69$ & $83.46$ & $69.95$ & $77.83$ & $63.91$ & $75.24$ \\
        De novo training  & Hybrid ($w_1=50$) &                        & $63.75$ & $82.19$ & $71.10$ & $77.98$ & $66.61$ & $76.83$ \\
        TL with GNN       & MSE               &                        & $65.28$ & $\mathbf{84.34}$ & $71.75$ & $\mathbf{79.66}$ & $66.93$ & $76.67$ \\
        TL with GNN       & Hybrid ($w_1=50$) &                        & $\mathbf{65.40}$ & $83.21$ & $\mathbf{72.38}$ & $79.51$ & $\mathbf{68.98}$ & $\mathbf{78.85}$ \\
        \midrule
        De novo training  & MSE               & \multirow{4}{*}{Large} & $66.74$ & $82.16$ & $70.88$ & $78.26$ & $68.58$ & $78.42$ \\
        De novo training  & Hybrid ($w_1=50$) &                        & $66.83$ & $81.63$ & $73.00$ & $79.29$ & $71.57$ & $80.50$ \\
        TL with GNN       & MSE               &                        & $\mathbf{69.10}$ & $\mathbf{84.30}$ & $73.47$ & $79.76$ & $72.28$ & $81.47$ \\
        TL with GNN       & Hybrid ($w_1=50$) &                        & $68.10$ & $83.21$ & $\mathbf{73.93}$ & $\mathbf{80.32}$ & $\mathbf{74.06}$ & $\mathbf{83.21}$ \\
        \bottomrule
    \end{tabular}
    \label{tab:mfspp_eval_cdf}
\end{table*}

\paragraph{Average performance}
Comparing the De novo training and TL with GNN methods, the latter outperforms the former in all metrics. For both small and large predictors, TL with GNN achieves lower average distance error, average room distance error, and average room rank, while also attaining higher average MIDR values compared to De novo training using the same loss function. This indicates that the TL with GNN approach effectively leverages the information embedded in the MIDR predictor's GNN module to enhance the performance of the MFSP predictor.

We compare the MSE loss with Hybrid loss within the TL with GNN method, using the Hybrid ($w_1=50$) case as an example. For the average MIDR metric, the Hybrid loss explicitly calculates MIDR, achieving a higher value than the MSE loss by up to $(2.46-2.40)\left/2.40\right.=2.50\%$ for the small predictor, with a smaller improvement for the large predictor. Conversely, as the MSE loss is computed based on the pseudo ground truth MFSP (similar to the distance error metric), it achieves a lower average distance error than the that of the Hybrid loss by up to $16\%$ for the large predictor. Interestingly, despite the average room distance error and room rank being calculated relative to the pseudo ground truth MFSP, the Hybrid loss performs similar to or even better than the MSE loss. This is likely attributed to the pseudo labels being accurate at the room-level granularity, even though the MFSP is not necessarily located at the center of the room. As a result, the MSE and Hybrid loss methods show varying performances across the distance error and room distance error metrics.

The Hybrid loss introduces a trade-off between achieving higher MIDR and lower distance error by adjusting the weights $w_1$ and $w_2$, which reflect the degree of reliance on the pseudo labels. In this analysis, the weight for the MSE term is fixed at $w_2=1$. By varying the weight for the MIDR term ($w_1$) from 10 to 100, the reliance on pseudo labels decreases as $w_1$ increases. This trade-off is evident when comparing results for $w_1=10$ and $w_1=50$, where $w_1=10$ results in about $10\%$ lower distance error but also yields a slightly lower MIDR, reduced by about $1\%$. When $w_1$ is set to a very large value, such as $w_1=100$, the reliance on the pseudo labels becomes minimal, and the difficulty of training an ``argmaxer'', as discussed in \secref{subsec:mfspp_pseudo_label}, negatively impacts the MFSP predictor's performance. For the average MIDR metric, the performance with $w_1=100$ is worse than that achieved with $w_1=50$. Although the average MIDR metric does not vary monotonically, it is noteworthy that it changes only slightly with $w_1$, indicating that $w_1$ is a robust choice.

Taking TL with GNN and Hybrid loss ($w_1=50$) as an example, the average room distance error reaches 1.15, i.e., the predicted MFSP is, on average, located within 1.15 rooms of the ground truth MFSP. Additionally, the average room rank is $6.73$ indicating that the predicted MFSP is approximately the 6th or 7th most fire-sensitive room among all rooms in the building, with each structure containing an average of over 87 rooms herein. These low values suggest that the MFSP predictor accurately identifies the MFSPs in the building.

\begin{figure*}[th!]
    \includegraphics[width=\linewidth]{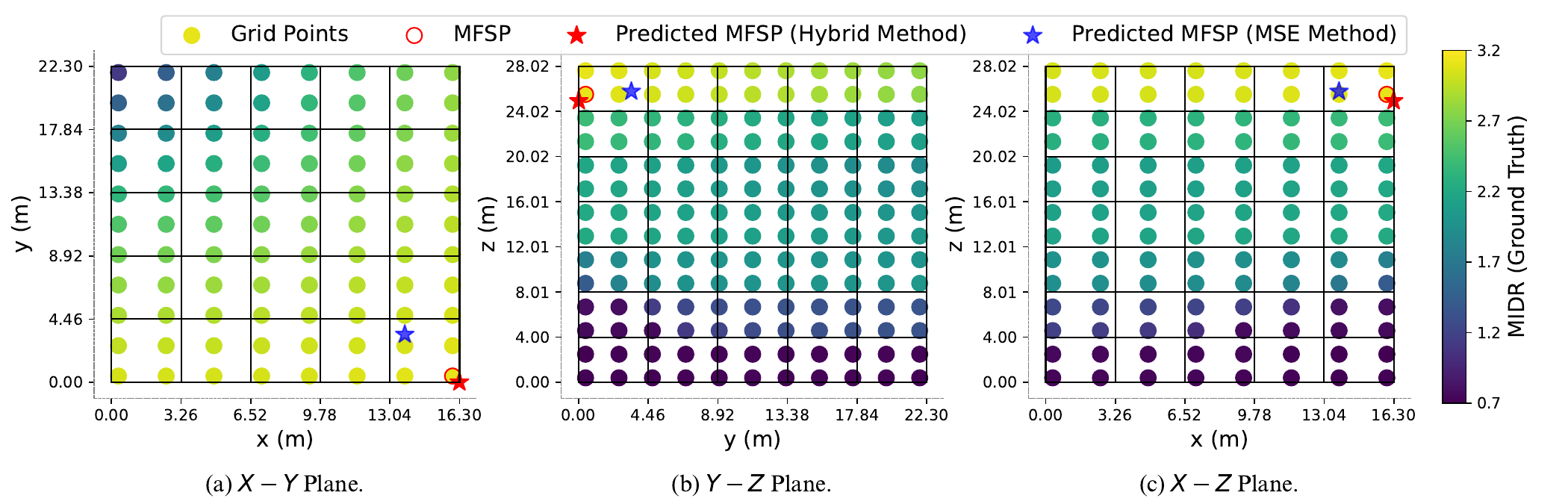}
    \caption{Illustrations of positions of ground truth and predicted MFSPs in three views of the building of  \texttt{ex2}.}
    \label{fig:eval_mfspp_cdf1} 
\end{figure*}

\paragraph{Quantile performance}
To further illustrate the performance of the MFSP predictor, we plotted the Cumulative Distribution Function (CDF), defined as ${\cal{F}}_{X}(a)={\cal{P}}(X \le a)$, where $\cal{P}$ indicates the probability and $X$ is a random variable representing the different metrics, i.e., distance error, room distance error, and room rank. Figures~\ref{fig:eval_mfspp_cdf}(a to c) display the results for the small predictor, while Figures~\ref{fig:eval_mfspp_cdf}(d to f) present the results for the large predictor. The key CDF values are summarized in \tabref{tab:mfspp_eval_cdf}. In \figref{fig:eval_mfspp_cdf}, the CDF curves of TL with GNN generally lie above those of De novo training, demonstrating that the former outperforms the latter in most cases. Besides, TL with GNN requires much less computational resources in training because it leverages a pre-trained GNN instead of training from scratch. The pre-trained model provides a well-initialized starting point, requiring only fine-tuning to adapt to the target task. This further highlights the efficiency and superiority of the approach of TL with GNN.

From \tabref{tab:mfspp_eval_cdf}, for the small network using TL with GNN and Hybird loss ($w_1=50$), $65.40\%$ of the cases achieve a distance error of less than $5$ meters. When the threshold is extended to $10$ meters, this ratio increases to $83.21\%$, surpassing the $82.19\%$ achieved with De novo training. Regarding room distance error, $\tilde{e}_{m}$, the large network using TL with GNN and Hybird loss achieves the highest probability of $73.93\%$ for $\tilde{e}_{m} \leq \sqrt{2}$, indicating that the predicted MFSP is either within the same room as the pseudo ground truth MFSP or in an adjacent room. Notably, for normalized rooms treated as unit-length squares (\secref{sec:7.2.1}), the longest distance between the centers of two diagonally adjacent rooms is $\sqrt{2}$. When the error threshold is further relaxed to $\tilde{e}_{m} \leq 2$, this probability increases to $80.32\%$, reflecting a high level of accuracy. For room rank, the large network employing TL with GNN and Hybird loss achieves the highest Top-5 accuracy, defined as the fraction of instances where the predicted MFSR ranks among the top 5 pseudo ground truth MFSRs, i.e., ${\cal{P}}\left(r_m \leq 5\right)$. This accuracy reaches $74.06\%$, representing an improvement of $2.49\%$ over De novo training. When using MSE loss, the TL with GNN method achieves a Top-5 accuracy of $72.28\%$, i.e., $3.7\%$ higher than that of De novo training. These results highlight the robustness of the TL with GNN approach, particularly when paired with Hybrid loss, in accurately identify the most fire-sensitive locations within buildings.

\paragraph{Case Study Building of \texttt{ex2}}

We take \texttt{ex2} (Figure~\ref{fig:example_generated_geometry}(b)) as an example to demonstrate the evaluation process of the MFSP prediction. In this case, the interior space of the building is discretized into a set of uniformly distributed grid points with approximately $\leq 2$ meters spacing in each direction, resulting in a total of $1,232$ grid points. We adopt a brute-force strategy by simulating each grid point as a potential fire point using the FEA simulation tool, and recording the corresponding MIDR for each grid point. Among all grid points, the one with the highest MIDR is considered the 3D vector of ground truth MFSP, denoted as $\text{MFSP}_{\gt} = [15.97, 0.45, 25.53]$ meters. This requires roughly $1,232 \times 48.34 \, \text{s}\approx 16.54$ hr to obtain this $\text{MFSP}_{\gt}$ (refer to \tabref{tab:midrp_time}). We then employ our proposed large MFSP predictor (TL with GNN) to predict the MFSP. Under the MSE-only training scheme, the predicted 3D vector of MFSP is $\text{MFSP}_{\text{mse}} = [13.73, 3.38, 25.78]$, resulting in a prediction error of $3.7$ meters. Under the Hybrid Loss ($w_1=50$), the predicted 3D vector of MFSP is $\text{MFSP}_{\text{hybrid}} = [16.29, 0.01, 24.95]$, with a much lower prediction error of $0.8$ meters. Since the MFSP predictors share the same architecture as the MIDR predictors, the inference time is still only several ms (refer to \tabref{tab:midrp_time}).

Figure~\ref{fig:eval_mfspp_cdf1} provides a visual comparison in three orthogonal planes ($X-Y$, $Y-Z$, and $X-Z$) containing $\text{MFSP}_{\gt}$. In each subplot, the MIDR values corresponding to all grid points are plotted as heatmaps, along with the projections of the predicted MFSPs. Notably, the grid point closest to the $\text{MFSP}_{\text{hybrid}}$ is exactly the same point as the $\text{MFSP}_{\gt}$, verifying the accuracy of the Hybrid loss method. From \figref{fig:eval_mfspp_cdf1}(a), we observe that the MIDR distribution in the $X-Y$ plane is clearly asymmetric. This is a deliberate consequence of introducing asymmetry into the simulation setup, in order to reflect the irregularity of real-world building geometries and uncertainty in the design procedures and building operations. Figures~\ref{fig:eval_mfspp_cdf1}(b) and (c) further illustrate the strong correlation between MIDR and the vertical position (floor level) of the fire point. The simulation results indicate that fires originating on lower floors tend to have limited impact on the overall structural stability in the relatively short simulation time. In contrast, fires occurring on higher floors are more likely to cause larger MIDRs in upper stories, posing a greater threat to the structural integrity. This insight suggests that structural strengthening for fire safety may need to be more focused on higher floors. Moreover, we note that the MIDR values do not exhibit a monotonic relationship with height. The MFSP does not occur at the highest elevation grid point, and this non-trivial characteristic is successfully captured by the MFSP predictor. The Hybrid loss method (indicated by the red star) achieves near-exact alignment with the ground truth MFSP, further validating its effectiveness.

In summary, the TL with GNN approach consistently outperforms De novo training across all evaluation metrics, highlighting the effectiveness of leveraging the MIDR predictor's GNN module to enhance the performance of the MFSP predictor. Two notable trade-offs arise when selecting the loss function. The first is between $w_1$ and $w_2$ within the Hybrid loss to include reliance  on the room-level granularity of the pseudo ground truth MFSP. The second is between MSE loss and Hybrid loss for higher efficiency and accuracy. While the Hybrid loss achieves slightly better results in most metrics (except for distance error), it incurs a higher computational cost due to the additional GNN forward pass required to calculate the MIDR loss and the backward pass through the MIDR predictor to calculate the gradient of the loss $L_{\text{MIDR}}$. In contrast, the MSE loss avoids this additional computational cost, making it more efficient but potentially less accurate in certain scenarios. This trade-off between computational cost and accuracy should be carefully considered based on the specific requirements and constraints of the application. If computational efficiency is a priority, the MSE loss may be more suitable. However, if higher accuracy in identifying sensitive fire points is critical, the Hybrid loss could provide a more robust solution.
\section{Limitations \& Future Work}
\label{sec:discussions}
\subsection{Performance Improvement} 
The proposed framework demonstrates promising results, yet there are opportunities to enhance its predictive capabilities. From a modeling perspective, the constraint of depth of GNN layers to match the number of building stories effectively mitigates over-smoothing but may limit the ability to model horizontal fire spreading in low-rise, large-area buildings. Future work could explore alternative architectures, e.g., introducing dummy nodes or improving message-passing mechanisms, to enhance cross-node communication in such scenarios. Additionally, the current implementation uses shallow MLPs for key components of the message-passing network, which may restrict the model's ability to capture complex interaction patterns. Exploring advanced GNN variants, e.g., GraphSAGE \cite{hamilton2017inductive} or Graph Attention Networks \cite{velickovic2018graph}, could improve prediction accuracy by better handling multi-scale spatial dependencies. A deeper analysis of the trade-off between MSE loss and MIDR loss could also further refine the model's performance.

Beyond a purely data-driven approach, integrating Physics-Informed Neural Networks (PINNs) could leverage physical laws to refine predictions. While PINNs have shown great promise in modeling physical systems, their effectiveness depends on well-formulated Partial Differential Equations (PDEs) ---a challenge in multi-physics problems like fire scenarios. In our case, the problem involves not only generating spatial thermal loads but also solving for the building's structural response under thermal loads. These complexities make the PDE formulation challenging, limiting the immediate application of PINNs. However, as hybrid methods and improved PDE modeling techniques evolve, incorporating PINNs into fire modeling remains a valuable research direction. Hybrid approaches that combine data-driven and physics-based methods could provide a more balanced and robust solution. 

Investigating the theoretical connections between GNN performance and FEA tools could strengthen the foundation for applying GNNs in structural analysis. The proposed framework is also compatible with other advanced ML techniques. For instance, ensemble learning \cite{alam_dynamic_2020} could integrate multimodal data, such as images or $\chi$ara scripts of the building structure, and self-supervised learning ---proven effective for sequential data like electroencephalograms \cite{rafiei_selfsupervised_2024a, rafiei_selfsupervised_2024}--- could offer insights into temporal thermal analysis. Furthermore, general techniques such as neural architecture search and hyperparameter optimization could  be incorporated to enhance both accuracy and practicality of the framework.

\subsection{Dataset Generation}
The lack of comprehensive datasets is a significant challenge in applying ML methods to fire-related research. Our work introduces a simplified rule-based approach for efficient fire dataset generation, but it may not fully capture the complexity of real-world structural geometries or thermal fields. Incorporating experimental measurements or advanced tools such as  the FDS could improve fidelity. 

Balancing fidelity, efficiency, and cost remains a key consideration. The proposed rule-based method could be refined to accommodate different structural configurations or fire patterns. For example, beyond tuning the parameters, as discussed in \secref{subsec:thermal_load_generation}, the ISO 834 curve could be replaced with a parametric time-temperature fire curve that accounts for additional factors such as ventilation conditions and fuel load density. In terms of modeling using FEA tools, some settings are chosen for simplicity of both modeling and simulation. The squared section, limited mid-element nodes, and introduced randomness of properties used in this work ensure the generation of a large dataset with acceptable cost to train the proposed predictors and test the workflow as a proof-of-concept. Such simplifications may introduce impractical scenarios in the generated dataset. Therefore, in the case when fidelity and accuracy involving practical cases are highly desired as in final designs (instead of the considered conceptual ones herein), detailed modeling and simulation with higher demand for computational resources can be applied. However, the the concept of MFSP and the workflow of its prediction are still applicable.

While physics-based simulations like OpenSees \cite{mckenna2011opensees, jiang_opensees_fire_2015} facilitates large-scale training data generation, their accuracy depends on assumptions about extreme load, e.g., fire, evolution and structural responses. Rigorous experimental validation and cross-comparison studies are essential to verify simulation fidelity, particularly for complex events like fires. Future research should prioritize such validation efforts to enhance confidence in simulation outputs and downstream predictive models. Additionally, addressing data scarcity through techniques such as TL, domain adaptation, or synthetic data augmentation could further improve model robustness. Leveraging pretrained models from this study as a foundation could enable adaptation to diverse structural configurations and fire scenarios.

\subsection{Beyond Structural Analysis}
The framework's differentiable agent and modular design offer transformative potential beyond fire scenario predictions, unlocking new applications across interdisciplinary domains. A key advantage of the differentiable agent is its ability to support training auxiliary predictors (e.g., ``argmaxers'') while also providing direct access to gradients that quantify how structural elements or environmental factors influence system-level responses. A natural extension of this work is structural analysis with combined loads, not only gravity and fire scenarios ---for instance, assessing the impact of both earthquakes and the subsequent fire-induced structural degradation. Beyond structural analysis, the framework's ability to identify worst-case scenarios can be generalized to other domains. For example, training a differentiable agent to simulate molecular dispersion (e.g., airborne viruses or chemical leaks) could identify the ``Most Diffusion-Sensitive Point'' (MDSP) ---the location where contaminants spread most rapidly. Such insight could inform critical design decisions, such as ensuring that chemical laboratories or quarantine rooms are positioned away from these MDSPs, or optimizing the ventilation systems to minimize cross-contamination risks.

The agent’s gradient information can directly map element-level properties (e.g., material stiffness, airflow resistance) to global performance metrics (e.g., structural drift, contaminant concentration). This enables physics-aware sensitivity analysis, where engineers can prioritize elements with the highest gradient magnitudes ---those that have the greatest impact on the system safety--- when making decisions about retrofitting, material selection, or structural requirements.

By addressing these limitations and exploring the outlined directions, future work can refine the proposed framework’s accuracy, expand its scope, and unlock new possibilities for data-driven structural engineering.
\section{Conclusions}
\label{sec:conclusion_future}

This study introduces a neural network-based framework for predicting the Most Fire-Sensitive Point (MFSP) in building structures, advancing fire safety assessment through a novel integration of data-driven and simulation-based approaches. The proposed framework leverages a differentiable agent model to efficiently simulate the outcome of the Finite Element Analysis (FEA) and effectively predict the Maximum Interstory Drift Ratio (MIDR) under fire conditions. This agent enables the rapid generation of MFSP-labeled data, directly supporting the training of an MFSP predictor. A rule-based temperature generation method was developed to facilitate the development of large-scale structural and fire simulation datasets and eventually forming a robust pipeline for data generation and neural network training. Key innovations of this work include: 
\begin{enumerate}
    \item The application of Graph Neural Networks (GNNs) to process structural data in fire scenarios.
    \item The integration of Transfer Learning (TL) to improve prediction performance.
    \item The introduction of an Edge Update (EU) mechanism to account for property changes in beams and columns during fire exposure. 
\end{enumerate}

The evaluation of the proposed framework demonstrates its effectiveness with the following two key points: 
\begin{enumerate}
\item The MIDR predictor achieved a maximum average Spearman's rank correlation coefficient of $0.74$, with an average correlation exceeding $0.91$ for severe fire scenarios.
\item The MFSP predictor, at the room level, attained Top-5 and Top-10 accuracies of up to $74.1\%$ and $83.2\%$, respectively.
\end{enumerate}
These results highlight the proposed framework potential to enhance building fire safety assessment by efficiently identifying critical structural vulnerabilities under fire conditions, particularly during the preliminary stage of the building design.

\section*{Acknowledgements}

The authors extend their heartfelt gratitude to Dr. M. Eslami, Mr. G. Zhou, Mr. C.M. Perez, Mr. W. Ruan, Mr. K.T. Pang, and Ms. L. Yang, members of \href{https://stairlab.berkeley.edu/}{STAIRlab}, for valuable discussions related to this research. Special thanks go to Prof. S.-L. Huang (Tsinghua University) and Prof. L. Zheng (MIT) for their insightful discussions that greatly enriched this study. We also deeply appreciate the Berkeley Research IT team for providing access to the Savio computing cluster, a critical resource for the ``STAIRlab'' project. This  high-performance computing infrastructure played a key role in data generation and network training, both of which were essential to the success of this research. Furthermore, we gratefully acknowledge the financial support provided to STAIRlab through the Taisei Chair in Civil Engineering at UC Berkeley. Both the computational and financial resources were indispensable in completing this work. Finally, we sincerely thank the editorial team for their efficient handling of the manuscript. We are particularly indebted to the nine anonymous peer reviewers, whose meticulous evaluations and thoughtful suggestions significantly improved the technical accuracy and overall clarity of this paper.

\bibliographystyle{plain}  
\bibliography{references}

\end{document}